\documentclass[11pt]{article}

\newif\ifarxiv
\arxivtrue
\ifarxiv
  \usepackage[preprint]{acl}
\else
  \usepackage[review]{acl}
\fi
\input{acl_lineno_fix}

\usepackage{times}
\usepackage{latexsym}
\usepackage[T1]{fontenc}
\usepackage[utf8]{inputenc}
\usepackage{microtype}
\usepackage{inconsolata}
\usepackage{url}
\usepackage{booktabs}
\usepackage{graphicx}
\usepackage{amsmath,amssymb}
\usepackage{multirow}
\usepackage{listings}
\usepackage{xcolor}
\usepackage{colortbl}
\usepackage{adjustbox}
\usepackage{wrapfig}
\usepackage{pifont}
\usepackage{xspace}
\usepackage{enumitem}
\usepackage{caption}

\usepackage{mdframed}
\usepackage{subcaption}

\usepackage[percent]{overpic}
\usepackage{pgfplots}
\pgfplotsset{compat=1.18}
\usetikzlibrary{patterns}
\usepgfplotslibrary{groupplots}
\pgfplotsset{
  scifig/.style={
    font=\footnotesize,
    tick label style={font=\footnotesize},
    label style={font=\footnotesize},
    title style={font=\footnotesize\bfseries, yshift=-3pt},
    legend style={font=\footnotesize, draw=none, fill=none, legend cell align=left,
      /tikz/every even column/.append style={column sep=3pt}},
    legend image code/.code={\draw[#1, draw=none] (0cm,-0.07cm) rectangle (0.22cm,0.14cm);},
    axis line style={-, oc-gray-6, line width=0.4pt},
    axis lines*=left,
    tick style={oc-gray-6, line width=0.4pt},
    ymajorgrids, grid style={oc-gray-2, line width=0.3pt},
    tick align=outside,
    xtick pos=left, ytick pos=left,
    enlarge x limits=0.15,
    clip=false,
    scaled y ticks=false,
  },
  tgt/.style={fill=targetfill, draw=targetdark, line width=0.3pt},
  inf/.style={fill=inferfill, draw=inferdark, line width=0.3pt, postaction={pattern=north west lines, pattern color=inferdark}},
  bth/.style={fill=bothfill, draw=bothdark, line width=0.3pt, postaction={pattern=crosshatch, pattern color=bothdark}},
  non/.style={fill=nonefill, draw=nonedark, line width=0.3pt, postaction={pattern=horizontal lines, pattern color=nonedark}},
  bas/.style={fill=basefill, draw=basedark, line width=0.3pt, postaction={pattern=crosshatch dots, pattern color=basedark}},
  cei/.style={fill=ceilfill, draw=ceildark, line width=0.3pt, postaction={pattern=grid, pattern color=white}},
  tgtmark/.style={mark=*, mark options={fill=targetfill, draw=targetdark, line width=0.3pt}},
  infmark/.style={mark=triangle*, mark options={fill=inferfill, draw=inferdark, line width=0.3pt}},
  bthmark/.style={mark=diamond*, mark options={fill=bothfill, draw=bothdark, line width=0.3pt}},
  nonmark/.style={mark=square*, mark options={fill=nonefill, draw=nonedark, line width=0.3pt}},
}

\newcommand{\frameworkName}{\textsc{SciSense}\xspace}
\newcommand{\datasetName}{\frameworkName-Traj\xspace}
\newcommand{\modelName}{\frameworkName-LM\xspace}

\newcommand{\Target}{\textcolor{targetdark}{\texttt{Hindsight}}\xspace}
\newcommand{\Infer}{\textcolor{inferdark}{\texttt{Foresight}}\xspace}
\newcommand{\Both}{\textcolor{bothdark}{\texttt{Both}}\xspace}
\newcommand{\None}{\textcolor{nonedark}{\texttt{None}}\xspace}

\newcommand{\cmark}{\textcolor{green!70!black}{\ding{51}}} 
\newcommand{\xmark}{\textcolor{red}{\ding{55}}} 
\newcommand{\pmark}{\textcolor{oc-orange-4}{\ding{115}}}

\newcommand{\ie}{i.e.,\xspace}

\newcommand{\styledtext}[3]{%
  \tcbox[
    on line,
    colback=#2,        
    colframe=#1,       
    coltext=#1,        
    boxsep=0pt,
    left=4pt, right=4pt,
    top=2pt, bottom=2pt,
    boxrule=1pt,       
    arc=4pt            
  ]{#3}%
}

\newcommand{\foraging}{\styledtext{oc-orange-8}{oc-orange-0}{\footnotesize\texttt{Foraging}\xspace}}
\newcommand{\shoebox}{\styledtext{oc-cyan-8}{oc-cyan-0}{\footnotesize\texttt{Shoebox}\xspace}}
\newcommand{\schema}{\styledtext{oc-violet-8}{oc-violet-0}{\footnotesize\texttt{Schema}\xspace}}
\newcommand{\hypothesis}{\styledtext{oc-pink-8}{oc-pink-0}{\footnotesize\texttt{Hypothesis}\xspace}}
\newcommand{\elaboration}{\styledtext{oc-yellow-8}{oc-yellow-0}{\footnotesize\texttt{Elaboration}\xspace}}
\newcommand{\questioning}{\styledtext{oc-red-8}{oc-red-0}{\footnotesize\texttt{Questioning}\xspace}}
\newcommand{\reframe}{\styledtext{oc-green-8}{oc-green-0}{\footnotesize\texttt{Reframe}\xspace}}
\newcommand{\presentation}{\styledtext{oc-indigo-8}{oc-indigo-0}{\footnotesize\texttt{Presentation}\xspace}}
\newcommand{\Tforag}{\foraging}
\newcommand{\Tshoebox}{\shoebox}
\newcommand{\Tschema}{\schema}
\newcommand{\Thyp}{\hypothesis}
\newcommand{\Telab}{\elaboration}
\newcommand{\Tquest}{\questioning}
\newcommand{\Treframe}{\reframe}
\newcommand{\Tpres}{\presentation}
\newcommand{\marginanno}[3]{%
  \noindent{\setlength{\fboxsep}{2pt}%
      \colorbox{black!8}{\parbox{\dimexpr\linewidth-2\fboxsep}{%
          \tiny\raggedright#1~\textbf{#2}\enspace#3}}}\par}

\usepackage[table]{xcolor}
\usepackage{xfp}

\newcommand{\heat}[3]{%
  \edef\pct{\fpeval{round(15 + 70*max(0,min(1,(#1-#2)/((#3-#2)+1e-9))))}}%
  \edef\heatcol{brown!\pct}%
  \expandafter\cellcolor\expandafter{\heatcol}{#1}%
}

\lstdefinestyle{paper-code}{
  basicstyle=\ttfamily\footnotesize,
  columns=fullflexible,
  breaklines=true,
  frame=single,
  rulecolor=\color{black!15},
  numbers=left,
  numberstyle=\tiny\color{black!50},
  xleftmargin=2em,
  framexleftmargin=1.5em,
  showstringspaces=false,
  tabsize=2
}

\usepackage[most]{tcolorbox}
\tcbset{
  colback=black!3,
  colframe=black!20,
  boxrule=0.4pt,
  arc=2pt,
  left=4pt,
  right=4pt,
  top=4pt,
  bottom=4pt,
}

\lstdefinelanguage{yaml}{
  sensitive=true,
  morecomment=[l]{\#},
  morestring=[b]",
  morestring=[b]',
  alsoletter={-},
  keywords={true,false,null,y,n,yes,no,on,off}
}

\definecolor{darkblue}{rgb}{0, 0, 0.5}

\definecolor{Red}{rgb}{0.768, 0.054, 0.054}
\definecolor{Blue}{rgb}{0.152, 0.294, 0.925}
\definecolor{Green}{rgb}{0,0.4,0.7}

\usepackage{cleveref}
\crefname{section}{§}{§§}

\title{Structure Liberates: How Constrained Sensemaking \\Produces More Novel Research Output}

\newcommand{\projectlinks}{}
\ifarxiv
\renewcommand{\projectlinks}{\\[0.5em]
  \raisebox{-0.4ex}{\includegraphics[height=1em]{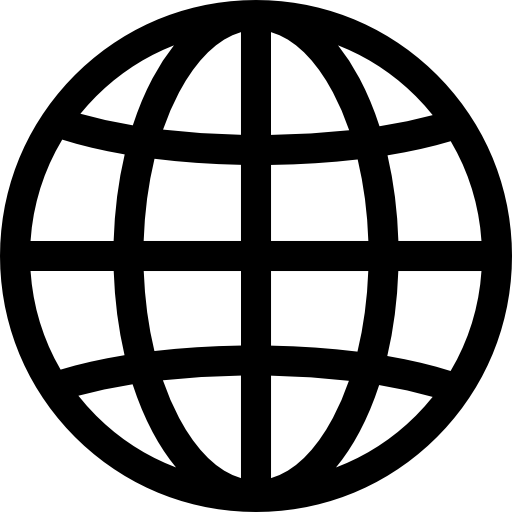}}\hspace{0.3em}\href{https://minnesotanlp.github.io/SciSense}{\texttt{Website}}
  \hspace{0.2cm}
  \raisebox{-0.4ex}{\includegraphics[height=1em]{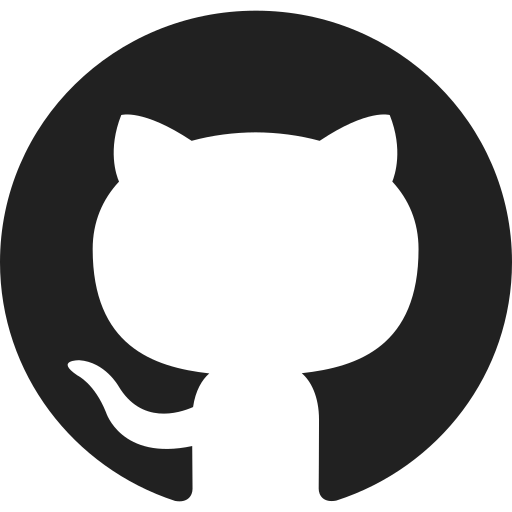}}\hspace{0.3em}\href{https://github.com/minnesotanlp/SciSense}{\texttt{Code}}
  \hspace{0.2cm}
  \raisebox{-0.4ex}{\includegraphics[height=1em]{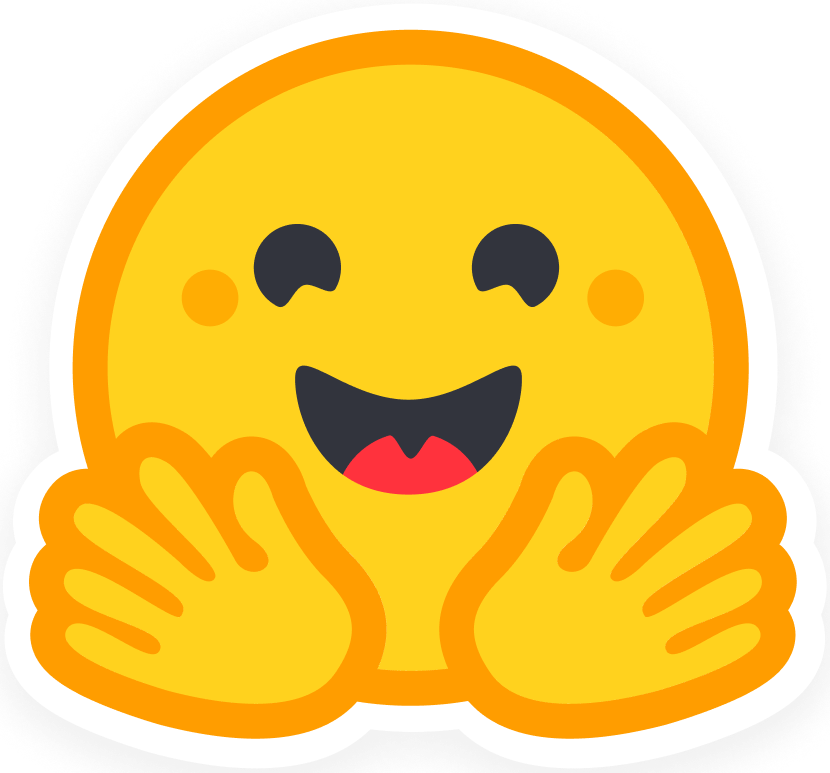}}\hspace{0.3em}\href{https://huggingface.co/collections/minnesotanlp/scisense}{\texttt{Models \& Dataset}}}
\fi
\author{James Mooney \quad Zae Myung Kim \quad Young-Jun Lee \quad Dongyeop Kang \\
  University of Minnesota\projectlinks}

\definecolor{infercol}{RGB}{80,200,120}
\definecolor{targetcol}{RGB}{100,100,220}
\definecolor{orangebg}{RGB}{255,245,230}
\definecolor{greenbg}{RGB}{230,248,230}
\definecolor{myorange}{RGB}{220,100,0}
\definecolor{mygreen}{RGB}{34,139,34}

\definecolor{oc-gray-0}{HTML}{F8F9FA}
\definecolor{oc-gray-1}{HTML}{F1F3F5}
\definecolor{oc-gray-2}{HTML}{E9ECEF}
\definecolor{oc-gray-3}{HTML}{DEE2E6}
\definecolor{oc-gray-4}{HTML}{CED4DA}
\definecolor{oc-gray-5}{HTML}{ADB5BD}
\definecolor{oc-gray-6}{HTML}{868E96}
\definecolor{oc-gray-7}{HTML}{495057}
\definecolor{oc-gray-8}{HTML}{343A40}
\definecolor{oc-gray-9}{HTML}{212529}

\definecolor{oc-red-0}{HTML}{FFF5F5}
\definecolor{oc-red-1}{HTML}{FFE3E3}
\definecolor{oc-red-2}{HTML}{FFC9C9}
\definecolor{oc-red-3}{HTML}{FFA8A8}
\definecolor{oc-red-4}{HTML}{FF8787}
\definecolor{oc-red-5}{HTML}{FF6B6B}
\definecolor{oc-red-6}{HTML}{FA5252}
\definecolor{oc-red-7}{HTML}{F03E3E}
\definecolor{oc-red-8}{HTML}{E03131}
\definecolor{oc-red-9}{HTML}{C92A2A}

\definecolor{oc-pink-0}{HTML}{FFF0F6}
\definecolor{oc-pink-1}{HTML}{FFDEEB}
\definecolor{oc-pink-2}{HTML}{FCC2D7}
\definecolor{oc-pink-3}{HTML}{FAA2C1}
\definecolor{oc-pink-4}{HTML}{F783AC}
\definecolor{oc-pink-5}{HTML}{F06595}
\definecolor{oc-pink-6}{HTML}{E64980}
\definecolor{oc-pink-7}{HTML}{D6336C}
\definecolor{oc-pink-8}{HTML}{C2255C}
\definecolor{oc-pink-9}{HTML}{A61E4D}

\definecolor{oc-grape-0}{HTML}{F8F0FC}
\definecolor{oc-grape-1}{HTML}{F3D9FA}
\definecolor{oc-grape-2}{HTML}{EEBEFA}
\definecolor{oc-grape-3}{HTML}{E599F7}
\definecolor{oc-grape-4}{HTML}{DA77F2}
\definecolor{oc-grape-5}{HTML}{CC5DE8}
\definecolor{oc-grape-6}{HTML}{BE4BDB}
\definecolor{oc-grape-7}{HTML}{AE3EC9}
\definecolor{oc-grape-8}{HTML}{9C36B5}
\definecolor{oc-grape-9}{HTML}{862E9C}

\definecolor{oc-violet-0}{HTML}{F3F0FF}
\definecolor{oc-violet-1}{HTML}{E5DBFF}
\definecolor{oc-violet-2}{HTML}{D0BFFF}
\definecolor{oc-violet-3}{HTML}{B197FC}
\definecolor{oc-violet-4}{HTML}{9775FA}
\definecolor{oc-violet-5}{HTML}{845EF7}
\definecolor{oc-violet-6}{HTML}{7950F2}
\definecolor{oc-violet-7}{HTML}{7048E8}
\definecolor{oc-violet-8}{HTML}{6741D9}
\definecolor{oc-violet-9}{HTML}{5F3DC4}

\definecolor{oc-indigo-0}{HTML}{EDF2FF}
\definecolor{oc-indigo-1}{HTML}{DBE4FF}
\definecolor{oc-indigo-2}{HTML}{BAC8FF}
\definecolor{oc-indigo-3}{HTML}{91A7FF}
\definecolor{oc-indigo-4}{HTML}{748FFC}
\definecolor{oc-indigo-5}{HTML}{5C7CFA}
\definecolor{oc-indigo-6}{HTML}{4C6EF5}
\definecolor{oc-indigo-7}{HTML}{4263EB}
\definecolor{oc-indigo-8}{HTML}{3B5BDB}
\definecolor{oc-indigo-9}{HTML}{364FC7}

\definecolor{oc-blue-0}{HTML}{E7F5FF}
\definecolor{oc-blue-1}{HTML}{D0EBFF}
\definecolor{oc-blue-2}{HTML}{A5D8FF}
\definecolor{oc-blue-3}{HTML}{74C0FC}
\definecolor{oc-blue-4}{HTML}{4DABF7}
\definecolor{oc-blue-5}{HTML}{339AF0}
\definecolor{oc-blue-6}{HTML}{228BE6}
\definecolor{oc-blue-7}{HTML}{1C7ED6}
\definecolor{oc-blue-8}{HTML}{1971C2}
\definecolor{oc-blue-9}{HTML}{1864AB}

\definecolor{oc-cyan-0}{HTML}{E3FAFC}
\definecolor{oc-cyan-1}{HTML}{C5F6FA}
\definecolor{oc-cyan-2}{HTML}{99E9F2}
\definecolor{oc-cyan-3}{HTML}{66D9E8}
\definecolor{oc-cyan-4}{HTML}{3BC9DB}
\definecolor{oc-cyan-5}{HTML}{22B8CF}
\definecolor{oc-cyan-6}{HTML}{15AABF}
\definecolor{oc-cyan-7}{HTML}{1098AD}
\definecolor{oc-cyan-8}{HTML}{0C8599}
\definecolor{oc-cyan-9}{HTML}{0B7285}

\definecolor{oc-teal-0}{HTML}{E6FCF5}
\definecolor{oc-teal-1}{HTML}{C3FAE8}
\definecolor{oc-teal-2}{HTML}{96F2D7}
\definecolor{oc-teal-3}{HTML}{63E6BE}
\definecolor{oc-teal-4}{HTML}{38D9A9}
\definecolor{oc-teal-5}{HTML}{20C997}
\definecolor{oc-teal-6}{HTML}{12B886}
\definecolor{oc-teal-7}{HTML}{0CA678}
\definecolor{oc-teal-8}{HTML}{099268}
\definecolor{oc-teal-9}{HTML}{087F5B}

\definecolor{oc-green-0}{HTML}{EBFBEE}
\definecolor{oc-green-1}{HTML}{D3F9D8}
\definecolor{oc-green-2}{HTML}{B2F2BB}
\definecolor{oc-green-3}{HTML}{8CE99A}
\definecolor{oc-green-4}{HTML}{69DB7C}
\definecolor{oc-green-5}{HTML}{51CF66}
\definecolor{oc-green-6}{HTML}{40C057}
\definecolor{oc-green-7}{HTML}{37B24D}
\definecolor{oc-green-8}{HTML}{2F9E44}
\definecolor{oc-green-9}{HTML}{2B8A3E}

\definecolor{oc-lime-0}{HTML}{F4FCE3}
\definecolor{oc-lime-1}{HTML}{E9FAC8}
\definecolor{oc-lime-2}{HTML}{D8F5A2}
\definecolor{oc-lime-3}{HTML}{C0EB75}
\definecolor{oc-lime-4}{HTML}{A9E34B}
\definecolor{oc-lime-5}{HTML}{94D82D}
\definecolor{oc-lime-6}{HTML}{82C91E}
\definecolor{oc-lime-7}{HTML}{74B816}
\definecolor{oc-lime-8}{HTML}{66A80F}
\definecolor{oc-lime-9}{HTML}{5C940D}

\definecolor{oc-yellow-0}{HTML}{FFF9DB}
\definecolor{oc-yellow-1}{HTML}{FFF3BF}
\definecolor{oc-yellow-2}{HTML}{FFEC99}
\definecolor{oc-yellow-3}{HTML}{FFE066}
\definecolor{oc-yellow-4}{HTML}{FFD43B}
\definecolor{oc-yellow-5}{HTML}{FCC419}
\definecolor{oc-yellow-6}{HTML}{FAB005}
\definecolor{oc-yellow-7}{HTML}{F59F00}
\definecolor{oc-yellow-8}{HTML}{F08C00}
\definecolor{oc-yellow-9}{HTML}{E67700}

\definecolor{oc-orange-0}{HTML}{FFF4E6}
\definecolor{oc-orange-1}{HTML}{FFE8CC}
\definecolor{oc-orange-2}{HTML}{FFD8A8}
\definecolor{oc-orange-3}{HTML}{FFC078}
\definecolor{oc-orange-4}{HTML}{FFA94D}
\definecolor{oc-orange-5}{HTML}{FF922B}
\definecolor{oc-orange-6}{HTML}{FD7E14}
\definecolor{oc-orange-7}{HTML}{F76707}
\definecolor{oc-orange-8}{HTML}{E8590C}
\definecolor{oc-orange-9}{HTML}{D9480F}

\definecolor{targetfill}{HTML}{FFA94D}   %
\definecolor{targetdark}{HTML}{D9480F}   %
\definecolor{inferfill}{HTML}{38D9A9}    %
\definecolor{inferdark}{HTML}{087F5B}    %
\definecolor{bothfill}{HTML}{74C0FC}     %
\definecolor{bothdark}{HTML}{1864AB}     %
\definecolor{nonefill}{HTML}{CED4DA}     %
\definecolor{nonedark}{HTML}{495057}     %
\definecolor{basefill}{HTML}{E9ECEF}     %
\definecolor{basedark}{HTML}{868E96}     %
\definecolor{ceilfill}{HTML}{343A40}     %
\definecolor{ceildark}{HTML}{212529}     %

\colorlet{stForaging}{oc-orange-8}
\colorlet{stShoebox}{oc-cyan-8}
\colorlet{stSchema}{oc-violet-8}
\colorlet{stHypothesis}{oc-pink-8}
\colorlet{stElaboration}{oc-yellow-8}
\colorlet{stQuestioning}{oc-red-8}
\colorlet{stReframe}{oc-green-8}
\colorlet{stPresentation}{oc-indigo-8}

\usepackage{stfloats}  %

\begin{document}

\maketitle

\begin{figure*}[b]
\centering
\begin{tikzpicture}
\begin{axis}[scifig, hide axis, scale only axis, width=1pt, height=1pt, xmin=0, xmax=1, ymin=0, ymax=1,
  legend columns=-1, legend style={at={(0.5,0.5)}, anchor=center, draw=none,
  /tikz/every even column/.append style={column sep=6pt}}]
\addlegendimage{bas}\addlegendentry{No-Plan baseline}
\addlegendimage{non}\addlegendentry{None}
\addlegendimage{inf}\addlegendentry{Foresight}
\addlegendimage{bth}\addlegendentry{Both}
\addlegendimage{tgt}\addlegendentry{Hindsight}
\end{axis}
\end{tikzpicture}\par\smallskip
\begin{subfigure}[b]{0.34\textwidth}
\centering
\begin{tikzpicture}
\begin{axis}[scifig, scale only axis, width=3.6cm, height=2.4cm,
ybar=0pt, bar width=4.6pt,
  ymin=0, ymax=0.30, ytick={0,0.1,0.2,0.3},
  yticklabel style={/pgf/number format/fixed, /pgf/number format/precision=1},
  ylabel={Diversity}, ylabel style={yshift=-3pt},
  symbolic x coords={Q4,G27,Q32,L70}, xtick=data,
  xticklabels={Q3-4B, G3-27B, Q3-32B, L3-70B},
  xticklabel style={font=\scriptsize, anchor=north, yshift=-1pt},
xlabel={Backbone}, xlabel style={yshift=5pt},
  enlarge x limits=0.16,
]
\addplot[non] coordinates {(Q4,0.074) (G27,0.082) (Q32,0.068) (L70,0.088)};
\addplot[inf] coordinates {(Q4,0.138) (G27,0.152) (Q32,0.128) (L70,0.162)};
\addplot[bth] coordinates {(Q4,0.172) (G27,0.184) (Q32,0.158) (L70,0.196)};
\addplot[tgt] coordinates {(Q4,0.238) (G27,0.252) (Q32,0.222) (L70,0.264)};
\end{axis}
\end{tikzpicture}
\subcaption{Diversity by backbone}\label{fig:overview_a}
\end{subfigure}\hfill
\begin{subfigure}[b]{0.31\textwidth}
\centering
\begin{tikzpicture}
\begin{axis}[scifig, scale only axis, width=3.4cm, height=2.4cm,
xmin=0.04, xmax=0.44, ymin=2.0, ymax=4.1,
  xtick={0.1,0.2,0.3,0.4}, ytick={2.5,3.0,3.5,4.0},
  xticklabel style={/pgf/number format/fixed, /pgf/number format/precision=1},
  yticklabel style={/pgf/number format/fixed, /pgf/number format/precision=1},
  xlabel={Diversity}, xlabel style={yshift=5pt},
  ylabel={Quality}, ylabel style={yshift=-3pt},
  xmajorgrids,
]
\addplot[only marks, mark=triangle*, mark size=2.1pt, mark options={fill=inferfill, draw=inferdark, line width=0.3pt, fill opacity=0.85}] coordinates {
(0.217,3.352)(0.123,3.273)(0.180,3.198)(0.141,3.177)(0.126,3.160)(0.082,3.155)(0.120,3.120)(0.092,3.112)(0.080,3.033)(0.182,2.976)(0.121,2.919)(0.230,2.919)(0.064,2.897)(0.108,2.836)(0.103,2.696)(0.253,2.608)(0.190,2.586)(0.160,2.578)(0.117,2.525)(0.137,2.521)(0.078,2.433)(0.091,2.429)(0.086,2.131)};
\addplot[only marks, mark=*, mark size=1.8pt, mark options={fill=targetfill, draw=targetdark, line width=0.3pt, fill opacity=0.85}] coordinates {
(0.332,3.930)(0.283,3.794)(0.304,3.790)(0.417,3.755)(0.242,3.707)(0.332,3.694)(0.281,3.645)(0.189,3.615)(0.279,3.589)(0.311,3.580)(0.317,3.571)(0.348,3.514)(0.334,3.453)(0.258,3.392)(0.145,3.374)(0.128,3.295)(0.161,3.287)(0.154,3.265)(0.181,3.260)(0.073,3.239)(0.156,3.204)(0.230,3.186)(0.253,3.155)(0.102,3.151)(0.153,3.028)};
\addplot[only marks, mark=triangle*, mark size=4.7pt, mark options={fill=inferdark, draw=white, line width=0.5pt}] coordinates {(0.134,2.854)};
\addplot[only marks, mark=*, mark size=3.9pt, mark options={fill=targetdark, draw=white, line width=0.5pt}] coordinates {(0.239,3.459)};
\draw[->, oc-gray-6, line width=0.5pt, shorten >=4pt, shorten <=4pt] (axis cs:0.134,2.854) -- (axis cs:0.239,3.459);
\node[font=\scriptsize, oc-gray-7, anchor=north west, inner sep=1pt] at (rel axis cs:0.02,0.98) {large = mean};
\end{axis}
\end{tikzpicture}
\subcaption{Quality vs.\ diversity (32B)}\label{fig:overview_b}
\end{subfigure}\hfill
\begin{subfigure}[b]{0.31\textwidth}
\centering
\begin{tikzpicture}
\begin{axis}[scifig, scale only axis, width=3.4cm, height=2.4cm,
ybar=0pt, bar width=4.2pt,
  ymin=0, ymax=1.0, ytick={0,0.25,0.5,0.75,1.0},
  yticklabel style={/pgf/number format/fixed, /pgf/number format/precision=2},
  ylabel={Artifact score}, ylabel style={yshift=-3pt},
  symbolic x coords={Exec,Ground,Util}, xtick=data,
  xticklabels={Exec., Grounding, Utility},
  xticklabel style={font=\footnotesize, anchor=north, yshift=-1pt},
  enlarge x limits=0.2,
  legend style={/tikz/every even column/.append style={column sep=6pt}},
xlabel={Criterion}, xlabel style={yshift=5pt},
]
\addplot[bas] coordinates {(Exec,0.61) (Ground,0.18) (Util,0.44)};
\addplot[non] coordinates {(Exec,0.79) (Ground,0.63) (Util,0.61)};
\addplot[inf] coordinates {(Exec,0.86) (Ground,0.70) (Util,0.67)};
\addplot[bth] coordinates {(Exec,0.88) (Ground,0.74) (Util,0.71)};
\addplot[tgt] coordinates {(Exec,0.91) (Ground,0.76) (Util,0.73)};
\end{axis}
\end{tikzpicture}
\subcaption{Downstream artifact scores}\label{fig:overview_c}
\end{subfigure}
\caption{%
Overview of findings for \modelName. Colors \emph{and} fill textures mark the condition throughout the paper (\Target solid, \Infer diagonal, \Both crosshatch, \None horizontal).
\textbf{(a)}~Diversity ($1-$mean pairwise similarity) for four backbones (Q3 = Qwen3, G3 = Gemma3, L3 = Llama3).
\textbf{(b)}~Per-citation-set quality vs.\ diversity (32B; \Target $\bullet$, \Infer $\blacktriangle$).
\textbf{(c)}~Scores of artifacts built by a coding agent from each trajectory type.
}
\label{fig:overview}
\end{figure*}

\begin{abstract}
Scientific discovery is an extended process of ideation---surveying prior work, forming hypotheses, and refining reasoning---yet existing LLM research agents treat this phase as a brief preamble despite its central role in research.
We introduce \frameworkName, a sensemaking-grounded framework that operationalizes ideation as a structured sequence of eight cognitive stages \citep{pirolli2005sensemaking}. We construct \datasetName, a dataset of 120K citation-conditioned research trajectories in two modes: \Target, where an LLM reconstructs the ideation path from a paper's cited works to the paper itself, and \Infer, where the LLM proposes novel directions from the same citations. We distill these into \modelName, a family of sensemaking LLMs spanning 3B to 70B parameters.
Contrary to the assumption that looser supervision promotes exploration, \Target-trained models improve trajectory quality by 2.0\% over \Infer-trained models while also producing more novel and diverse outputs, across all eight model families we test. The advantage propagates downstream: coding agents conditioned on \Target trajectories produce research artifacts with higher executability, scientific grounding, and utility than those conditioned on \Infer trajectories.
This suggests that outcome-anchored ideation reduces the cognitive burden on downstream agents, freeing them to explore more creatively. \frameworkName offers both a practical tool for augmenting LLM-driven research workflows and a principled testbed for studying how planning shapes scientific discovery.
\end{abstract}

\begin{figure*}[t]
    \centering 
    \adjustbox{trim=0pt {.632\height} 0pt 0pt, clip}{\includegraphics[
        width=\textwidth,
        trim={0 3cm 0 3cm},
        clip
    ]{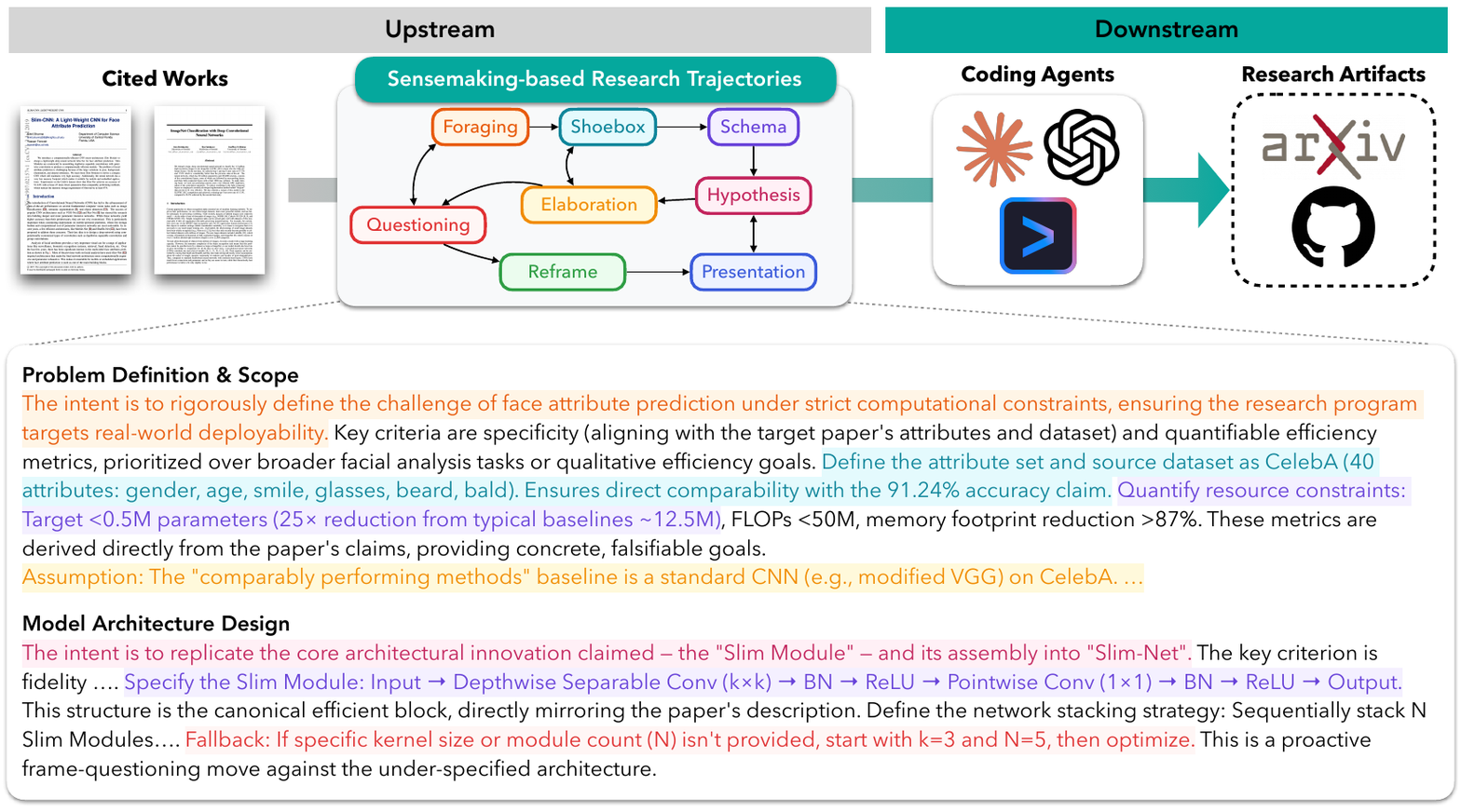}}
\caption{\frameworkName inserts an explicit \emph{sensemaking} phase between a paper's cited works and the downstream coding agents that produce code and paper artifacts. Instead of a one-paragraph plan, the model writes a structured research trajectory that moves through eight sensemaking stages (Table~\ref{main_tab:sensemaking_stages}); this trajectory then conditions the coding agent. Example trajectories are shown in Figure~\ref{fig:framework-comparison}.}
\label{fig:motivation}

\end{figure*}

\section{Introduction}
\label{sec:intro}

Scientific discovery is an iterative refinement loop: researchers survey prior work, form hypotheses, refine their reasoning, and only then implement. Recently, LLM-driven research agents~\citep{lu2024ai,ai_coscientist_arxiv_2025,yamada2025ai,baek2025researchagent,seo2025paper2code} have made substantial progress on scientific discovery tasks. These agents typically operate in two phases: (1) an \textbf{upstream} phase of ideation and planning, in which novel ideas are proposed and hypotheses formulated, and (2) a \textbf{downstream} phase, in which those ideas are realized through code implementation, experimental execution, and result analysis. Whatever happens upstream propagates downstream.

Yet existing agents treat the upstream phase as a brief preamble: a paragraph or two of abstract statements. The upstream process of human researchers is far richer, unfolding over many structured steps. This gap between the simplified planning in current systems and the structured process followed by human researchers is the focus of this paper.
To close it, we reinterpret the upstream phase through the \textbf{sensemaking} framework~\citep{pirolli2005sensemaking}, which describes how researchers structure information into explanatory frameworks, and operationalize it as eight cognitive stages (Tab~\ref{main_tab:sensemaking_stages}).

Building on sensemaking principles, we introduce \frameworkName, a sensemaking-grounded framework for scientific discovery with end-to-end support. We construct \datasetName, a 120K-trajectory dataset of sensemaking-based research trajectories conditioned on \textit{citation neighborhoods}, in two complementary modes: \Target, where an LLM reconstructs a plausible ideation trajectory leading from the cited works to a known paper, and \Infer, where the LLM proposes novel research directions grounded in the same citations. Using this dataset, we distill sensemaking into \modelName, a family of LLMs ranging from 3B to 70B parameters.

Figure~\ref{fig:overview} summarizes our central empirical findings. Contrary to the assumption that weaker constraints promote broader exploration, outcome-anchored \Target supervision yields trajectories that are simultaneously \emph{more diverse} and \emph{higher quality} than open-ended \Infer supervision, across all evaluated model families and scales. \Target-trained models are the most diverse under every automatic metric and gain 2.0\% in aggregate plan quality over \Infer---there is no quality--diversity trade-off. Mechanistically, \Target plans emphasize the constructive stages (\textit{Schema}, \textit{Elaboration}), whereas \Infer plans front-load \textit{Hypothesis} and \textit{Questioning}. The advantage transfers downstream: when a coding agent is conditioned on \Target trajectories, the resulting research artifacts improve in scientific grounding, executability, and downstream utility.

\begin{figure*}[t]
    \centering
    \definecolor{bandHind}{HTML}{F0997A}\definecolor{bandFore}{HTML}{7AC7C4}%
    \begin{tikzpicture}
      \node[anchor=north west, inner sep=0pt] at (0,0) {\adjustbox{trim=0pt 0pt 0pt {.175\height}, clip}{\includegraphics[width=\textwidth]{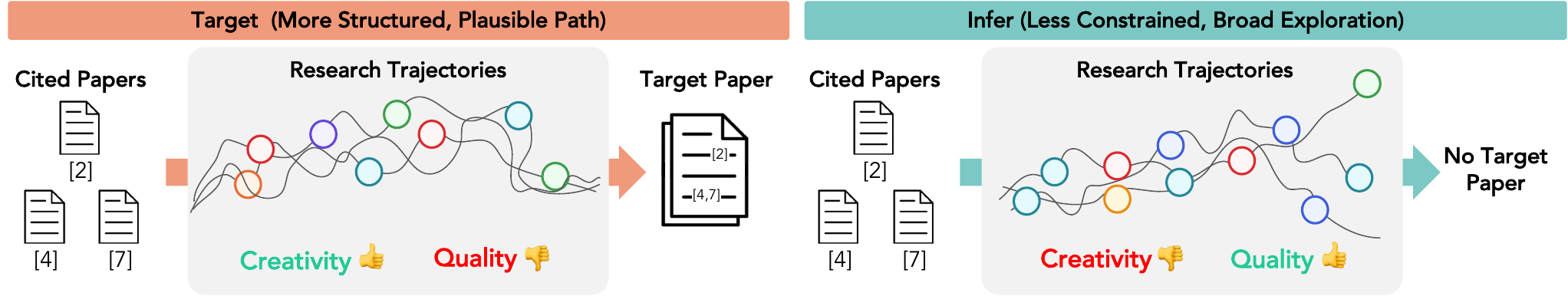}}};
      \fill[bandHind] (0,0.06cm) rectangle (0.5\textwidth-0.04cm,0.47cm);
      \fill[bandFore] (0.5\textwidth+0.06cm,0.06cm) rectangle (\textwidth,0.47cm);
      \node[font=\sffamily\fontsize{6.5}{7}\selectfont, text=black!85] at (0.25\textwidth,0.265cm) {\textbf{Hindsight} (More Structured, Plausible Path)};
      \node[font=\sffamily\fontsize{6.5}{7}\selectfont, text=black!85] at (0.75\textwidth,0.265cm) {\textbf{Foresight} (Less Constrained, Broad Exploration)};
    \end{tikzpicture}
\caption{The two trajectory modes in \datasetName. \Target trajectories (left) are anchored to a known outcome: the teacher reconstructs the sensemaking path from the cited papers to the target paper. \Infer trajectories (right) have no such anchor and propose forward directions from the same citations. One might expect the anchored mode to trade diversity for quality; Section~\ref{main_sec:experiments} shows that \Target wins on \emph{both}.}
\label{fig:infer_target_high_level}
\end{figure*}

Practitioners can use \frameworkName as a testbed for improving the ideation and planning capabilities of LLM-driven research agents. Our results suggest that reconstructive supervision yields higher-quality \emph{and} more diverse ideation, and that this structure enhances the creativity of downstream coding agents---underscoring the value of principled, structured planning. Overall, \frameworkName advances our understanding of how upstream planning shapes scientific discovery and provides a principled framework for developing and evaluating next-generation research agents.

\section{\frameworkName Framework}
\label{sec:method}

\frameworkName is the first sensemaking-based framework for automated scientific discovery that provides end-to-end support---from citation-grounded corpus construction to scalable trajectory learning and downstream evaluation---while remaining fully open-source (Table~\ref{tab:framework_comparison}).
The framework operates in three stages---corpus construction, trajectory generation, and model distillation---each grounded in sensemaking principles (Figure~\ref{fig:motivation}).
\cref{main_sec:preliminaries} introduces the sensemaking principle; \cref{main_sec:problem_formulation} formalises the discovery problem; \cref{main_sec:dataset_construction} describes dataset construction; \cref{main_sec:model_training} covers model training; and \cref{main_sec:experiments} defines the evaluation protocol.

\subsection{Preliminaries: Sensemaking Principle}
\label{main_sec:preliminaries}

The \emph{sensemaking principle}~\citep{pirolli2005sensemaking} describes how researchers
iteratively construct meaning from prior work---collecting, organising, and synthesising evidence
into coherent, testable, and procedurally grounded research trajectories that persuade others
through empirical justification.
We operationalise sensemaking as an ordered sequence of eight stages (Table~\ref{main_tab:sensemaking_stages}) that mirror how a researcher turns a body of literature into a research contribution.

\begin{table}[t]
\renewcommand{\arraystretch}{0.6}
  \centering
  \footnotesize

\begin{tabular}{r l}
\toprule
\foraging & Retrieve prior work and gaps \\
    \shoebox      & Organise extracted evidence \\
    \schema       & Form initial problem framing \\
    \hypothesis   & Define testable hypotheses \\
    \elaboration  & Develop methodological plan \\
    \questioning  & Assess feasibility and novelty \\
    \reframe      & Revise formulation or approach \\
    \presentation & Synthesize final trajectory \\
    \bottomrule
  \end{tabular}

\caption{The eight sensemaking stages, in order. Stage labels are reused in the trajectory excerpts.}
\label{main_tab:sensemaking_stages}
\end{table}

\subsection{Problem Formulation for Automated Scientific Discovery in \frameworkName}
\label{main_sec:problem_formulation}

The goal of automated scientific discovery is to enable an LLM $\mathcal{M}$ to autonomously
conduct research by following a scientific pipeline $\mathcal{P}$, analogous to human researchers.
Following prior work~\citep{lu2024ai,yamada2025ai}, this pipeline consists of four stages:
ideation and planning ($\mathcal{P}_{\text{plan}}$), experimentation ($\mathcal{P}_{\text{expr}}$),
analysis ($\mathcal{P}_{\text{analysis}}$), and paper writing ($\mathcal{P}_{\text{write}}$).
Given contextual information $\mathcal{C} = \{x_\texttt{paper}, x_\texttt{code}\}$, the pipeline
produces outputs $O := \mathcal{P}(\mathcal{C};\mathcal{M})$, where $O = \{y_\texttt{paper},
y_\texttt{code}\}$.
The standard pipeline is formalised as:
\begin{align*}
P &= \colorbox{gray!20}{$\mathcal{P}_{\text{plan}}(\mathcal{C})$}, &
E &= \mathcal{P}_{\text{expr}}(\mathcal{C}, P), \\
A &= \mathcal{P}_{\text{analysis}}(\mathcal{C}, P, E), &
O &= \mathcal{P}_{\text{write}}(\mathcal{C}, P, E, A).
\end{align*}

The planning stage $\mathcal{P}_{\text{plan}}$ is the critical bottleneck: shallow plans lead to incoherent or template-like downstream outputs.
\frameworkName addresses this by replacing the shallow plan $P$ with a sensemaking trajectory $S$, which encodes structured, fine-grained procedural knowledge about the research process.
Let $\mathcal{C}_{\text{cited}}(x_{\texttt{paper}}) = \{c_1, ..., c_N\}$ denote the citation neighborhood of a target paper $x_{\texttt{paper}}$, where $N = \left|\mathcal{C}_{\text{cited}}(x_{\texttt{paper}})\right|$. We define the extended context as $\mathcal{C}' = \mathcal{C} \cup \mathcal{C}_{\text{cited}}(x_{\texttt{paper}})$.
The \frameworkName pipeline is then:
\begin{align*}
S &= \colorbox{oc-teal-2}{$\mathcal{P}_{\text{sense}}(\mathcal{C}_\text{cited})$}, &
E &= \mathcal{P}_{\text{expr}}(\mathcal{C}', S), \\
A &= \mathcal{P}_{\text{analysis}}(\mathcal{C}', S, E), &
O &= \mathcal{P}_{\text{write}}(\mathcal{C}', S, E, A).
\end{align*}

The key difference is that $\mathcal{P}_{\text{sense}}$ conditions exclusively on the cited
works---not on the target paper itself---mimicking the information available to a researcher
\emph{before} a paper is written.

\subsection{\datasetName: A Collection of Sensemaking-Based Research Trajectories}
\label{main_sec:dataset_construction}

We construct \datasetName, a large-scale collection of sensemaking-based research trajectories, comprising two trajectory sets: \Target and \Infer. In all steps of our dataset construction process, we use \texttt{Qwen3-235B-A22B-Instruct-2507}~\citep{yang2025qwen3technicalreport} as our LLM. All prompt templates used in our dataset construction are presented in Appendix~\ref{app:pipeline}.

\textbf{Motivation for Sensemaking-Grounded Corpus Design.}
We design \datasetName{} to mirror the information structure of real scientific discovery, where researchers reason from a citation neighbourhood rather than from a completed paper. Accordingly, trajectories are grounded in cited works and structured using the eight-stage sensemaking process, which captures how evidence is collected, organized, and transformed into a research contribution. 

\textbf{Step 1: Corpus and Citation Neighbourhood Collection.}
We begin by collecting a large-scale academic text corpus from the Semantic Scholar Open Research Corpus (S2ORC)~\citep{s2orc}. Each instance in S2ORC consists of a target paper $x_\texttt{paper}$, its abstract, and rich paper-level metadata, including citation graphs, section boundaries, and inline citation markers. For each target paper $x_\texttt{paper}$, we extract the set of all papers it cites, \ie its \textit{citation neighbourhood} $\mathcal{C}_\text{cited}(x_\texttt{paper})$, which we use for research trajectory generation.

\textbf{Step 2: Cited Paper Summary Generation.}
We prompt an LLM to generate a concise summary $s_i$ for each paper $c_i \in \mathcal{C}_{\text{cited}}$, capturing its core contributions, methods, and key empirical findings. In addition, we generate a concise summary $s_x$ for the target paper $x_\texttt{paper}$, which is later used to construct \Target. Note that these generated summaries are used solely for dataset construction and are not utilized during training or evaluation.

\textbf{Step 3: Sensemaking-based Research Trajectory Generation.}
Given the generated summaries of cited papers $s_i$, we prompt the LLM to produce structured research trajectories that follow the sensemaking principle (in \cref{main_sec:preliminaries}), under two distinct settings:

\begin{itemize}[leftmargin=1em, itemsep=2pt, parsep=0pt, topsep=4pt]

    \item \textbf{\Target} (reasoning \emph{in hindsight}, the outcome is known)\textbf{:} Given the summaries of the cited works (\ie $s_1, \ldots, s_N$) and the summary of the target paper $s_x$, we prompt the LLM to reconstruct a principled research process that connects the cited works to the target paper. Since these trajectories are grounded in a known outcome (\ie the target paper), models trained on such trajectories can learn to reconstruct the author's reasoning process with high fidelity.

    \item \textbf{\Infer} (reasoning \emph{in foresight}, the outcome is unknown)\textbf{:} Given the summaries of the cited works (\ie $s_1, \ldots, s_N$), we prompt the LLM to propose novel research directions grounded in the citation neighbourhood $\mathcal{C}_\text{cited}$. These trajectories are designed to capture open-ended ideation, reflecting the stage at which researchers explore diverse possibilities before converging on a specific research direction; accordingly, models trained on such trajectories are expected to explore more diverse research processes.
    
    \item \textbf{\Both:} This setting includes both \Target and \Infer, enabling the model to learn both reconstructive reasoning (\ie how the target paper emerges from its cited works) and generative reasoning (\ie how the target paper can inspire future research directions).

\end{itemize}

All these trajectories are represented as markdown-style outputs to ensure structural consistency and alignment with the sensemaking framework.

\subsection{\modelName: Distilling Sensemaking into Smaller Language Models}
\label{main_sec:model_training}

To induce a ``sensemaking procedure'' in LLMs, we introduce a new family of LLMs that generate sensemaking-based research trajectories given only the raw text of cited papers, spanning a range of model sizes: \modelName-$\{3, 4, 27, 30, 32, 70\}$B. Specifically, we fine-tune multiple LLM backbones, including Qwen3~\citep{yang2025qwen3technicalreport}, Gemma-3~\citep{gemmateam2025gemma3technicalreport}, and LLaMA~3~\citep{llama3modelcard}, covering both base and instruction-tuned variants.  
During training, we provide truncated\footnote{To fit the model's context size, we adopt different truncation strategies (Appendix~\ref{app:sft}).} text of the cited papers along with sensemaking-based, trajectory-specific instructions as input, and the model is trained to sequentially generate the corresponding sensemaking-based research trajectory.
Crucially, the target paper is removed entirely from the distillation inputs, even for \Target trajectories: at inference time no such supervisory signal exists, so the student only ever sees cited papers. This design lets us ask two questions in our evaluation: whether the sensemaking procedure itself can be distilled into smaller language models, and whether anchoring the \emph{teacher} to a target paper (but never the student) still changes what the student learns, \ie the difference between \Target- and \Infer-trained models.

\section{Experiments}
\label{main_sec:experiments}

We evaluate every trajectory in two complementary setups (the full pipeline is illustrated in Figure~\ref{fig:eval_pipeline}, Appendix~\ref{app:eval_protocol}):
\begin{itemize}[leftmargin=*]
\item \textbf{Upstream (trajectory-level) evaluation} scores the trajectory itself, before anything is executed. An LLM judge rates its \emph{quality} on an eight-dimension rubric, validated against three human annotators (Section~\ref{sec:quality}), and we measure its \emph{diversity} over five samples per citation set with four similarity metrics (Section~\ref{sec:diversity}).
\item \textbf{Downstream (artifact-level) evaluation} asks whether the trajectory is useful. A frontier coding agent receives the trajectory and the same cited works and must produce a runnable repository and a short paper; a judge from a different model family scores these artifacts on executability, scientific grounding, and utility (Section~\ref{sec:downstream_eval}).
\end{itemize}

\begin{table*}[t]
\centering
\small
\begin{minipage}[b]{0.46\textwidth}
\centering
\setlength{\tabcolsep}{5pt}
\begin{tabular}{lccc}
\toprule
\textbf{Split} & \textbf{\Infer} & \textbf{\Target} & \textbf{Total} \\
\midrule
Train & 50K & 50K & 100K \\
Validation & 5K & 5K & 10K \\
Test & 5K & 5K & 10K \\
\midrule
Total & 60K & 60K & 120K \\
Avg.\ length (tokens) & 2031 & 2918 & 2475 \\
Avg.\ papers cited & 5.32 & 4.39 & 4.86 \\
Avg.\ neighb.\ size & -- & -- & 7.78 \\
\bottomrule
\end{tabular}
\end{minipage}\hfill
\begin{minipage}[b]{0.50\textwidth}
\centering
\setlength{\tabcolsep}{4pt}
\begin{tabular}{lccccc}
\toprule
\textbf{Annotators} & Nov. & Sig. & Grnd. & Snd. & Meth. \\
\midrule
Humans (3) & 0.684 & 0.749 & 0.548 & 0.728 & 0.752 \\
Humans + judge & 0.656 & 0.766 & 0.507 & 0.716 & 0.773 \\
\midrule
 & Feas. & Sense. & Clar. & & \textbf{Overall} \\
\midrule
Humans (3) & 0.796 & 0.612 & 0.736 & & \textbf{0.701} \\
Humans + judge & 0.763 & 0.594 & 0.745 & & \textbf{0.690} \\
\bottomrule
\end{tabular}
\end{minipage}

\begin{minipage}[t]{0.46\textwidth}
\caption{\datasetName statistics. \emph{Papers cited}: references used inside a trajectory; \emph{neighb.\ size}: cited papers available as input.}
\label{tab:framework_dataset_stats_main}
\end{minipage}\hfill
\begin{minipage}[t]{0.50\textwidth}
\caption{Inter-annotator agreement (Krippendorff's $\alpha$) per rubric dimension on 100 trajectories, among the three human annotators alone and with the LLM judge added as a fourth annotator. Agreement is essentially unchanged, so the judge is used for all scaled evaluations.}
\label{tab:krippendorff_alpha}
\end{minipage}
\end{table*}

We investigate the following research questions: \textbf{RQ1. (Section \ref{sec:quality})} Does grounding in sensemaking principles improve the quality of generated research trajectories? \textbf{RQ2. (Section \ref{sec:diversity})} Do open-ended, ideation-driven trajectories yield more diverse research directions than reconstruction-oriented trajectories? \textbf{RQ3. (Section \ref{sec:downstream_eval})} Do sensemaking-based research trajectories enable coding agents to produce higher-quality downstream artifacts?

\subsection{Experimental Setup}
\datasetName contains 120{,}000 citation-conditioned trajectories, evenly split between \Infer and \Target, with matched 50K/5K/5K train/validation/test partitions per condition (Table~\ref{tab:framework_dataset_stats_main}).
We distill the trajectories into eight open-weight backbones spanning the Qwen3, Gemma-3, and Llama-3 families (Table~\ref{tab:model_configs}). We compare four supervision conditions throughout: \Target, \Infer, \Both (trained on the union), and \None, the undistilled checkpoint without any trajectory supervision, which serves as our baseline.
Training details are in Appendix~\ref{app:sft}.

\subsection{RQ1: Producing Higher Quality Research Plans}
\label{sec:quality}

\textbf{Manual Annotations.}
LLM judges are known to exhibit systematic biases \citep{koo2024cobbler}, so before scaling up with an LLM judge we validate it against human annotation. Three graduate-level annotators with prior experience evaluating LLM outputs independently reviewed a balanced sample of 100 generated trajectories drawn uniformly across the four conditions (\None, \Infer, \Both, \Target) and across distilled models. Annotators received the same written instructions and rubric as the LLM judge (\texttt{Qwen3.5-35B-A3B-FP8}; full prompt in Appendix~\ref{app:llm_as_judge}). The rubric scores eight dimensions on a 1--5 scale: Novelty, Significance, Grounding, Soundness, Methodological Rigor, Feasibility, Sensemaking, and Clarity.

\begin{figure}[t]
\centering
\begin{tikzpicture}
\begin{axis}[
  scifig,
  width=\linewidth, height=4.1cm,
  ybar=0pt, bar width=3.2pt,
  ymin=0, ymax=0.18, ytick={0,0.05,0.10,0.15},
  yticklabel style={/pgf/number format/fixed, /pgf/number format/precision=2},
  ylabel={$\Delta$ judge score vs.\ None}, ylabel style={yshift=-4pt},
  symbolic x coords={Nov,Sig,Grnd,Snd,Meth,Feas,Sns,Clar,Ovr},
  xtick=data,
  xticklabels={Nov., Sig., Grnd., Snd., Meth., Feas., Sense., Clar., \textbf{Overall}},
  xticklabel style={font=\footnotesize, rotate=35, anchor=north east, xshift=3pt, yshift=-1pt},
  enlarge x limits=0.07,
  legend columns=3, legend style={at={(0.5,1.02)}, anchor=south, column sep=3pt, font=\footnotesize},
]
\addplot[inf] coordinates {(Nov,0.04) (Sig,0.04) (Grnd,0.16) (Snd,0.05) (Meth,0.06) (Feas,0.05) (Sns,0.04) (Clar,0.04) (Ovr,0.06)};
\addplot[bth] coordinates {(Nov,0.06) (Sig,0.06) (Grnd,0.09) (Snd,0.08) (Meth,0.11) (Feas,0.09) (Sns,0.08) (Clar,0.11) (Ovr,0.09)};
\addplot[tgt] coordinates {(Nov,0.13) (Sig,0.13) (Grnd,0.07) (Snd,0.10) (Meth,0.12) (Feas,0.11) (Sns,0.10) (Clar,0.12) (Ovr,0.11)};
\legend{Foresight, Both, Hindsight}
\draw[oc-gray-5, dashed, line width=0.4pt] ([xshift=-13pt]axis cs:Ovr,0) -- ([xshift=-13pt]axis cs:Ovr,0.18);
\end{axis}
\end{tikzpicture}
\caption{LLM-judge gains over the \None checkpoint for \modelName-32B, per rubric dimension. \Target improves most on nearly every dimension and overall; \Infer stands out only on \emph{Grounding} in prior work. Absolute scores for all model families are in Tab~\ref{tab:rubric_scores_models_extended}.}
\label{fig:q32-rubric}
\end{figure}

Table~\ref{tab:krippendorff_alpha} reports inter-annotator agreement (Krippendorff's $\alpha$) per rubric dimension. Agreement is substantial among the human annotators ($\alpha=0.70$ overall) and barely changes when the LLM judge is added as a fourth annotator ($\alpha=0.69$), supporting its use as a proxy evaluator in the rest of this section.

\paragraph{Scaled Evaluations.}
With the judge validated, we score all distilled models on 1K held-out citation sets each (full results in Appendix~\ref{app:llm_as_judge}). Figure~\ref{fig:q32-rubric} shows the gains over the \None checkpoint for \modelName-32B. \Target yields the highest overall quality and the largest gains on most dimensions, whereas \Infer yields the largest gain on \textit{Grounding}, \ie how well the trajectory ties its claims to the cited works. The same ordering holds across all eight model families (Table~\ref{tab:rubric_scores_models_extended}).

\paragraph{RL Post-Training.} Applying GRPO post-training to the \Both checkpoint of \modelName-32B, with the judge rubric as the reward, yields modest but consistent gains on all eight rubric dimensions (Overall: 2.51\,$\to$\,2.53); the 4B-Instruct \Both checkpoint shows the same pattern (2.40\,$\to$\,2.42). Implementation details and per-dimension results are in Appendix~\ref{app:rl_training}.

\subsection{RQ2: Scientific Creativity of Generated Research Trajectories}
\label{sec:diversity}

Creative research trajectories must be both diverse \emph{and} high quality. The conventional expectation is that anchoring supervision to a known outcome (\Target) should \emph{reduce} output diversity relative to open-ended ideation (\Infer), although prior work shows that how generation is structured can itself raise the diversity of LLM outputs \citep{hayati2024diverse}. We test this with four complementary diversity metrics computed over repeated sampling: each model generates five trajectories for each of 100 held-out citation sets from the test partition.
\begin{figure}[t]
\centering
\begin{tikzpicture}
\begin{axis}[
  scifig,
font=\small, tick label style={font=\small}, label style={font=\small},
  width=0.95\linewidth, height=5.4cm,
  xmin=0.0, xmax=0.30, ymin=2.38, ymax=2.56,
  xtick={0,0.1,0.2,0.3}, ytick={2.40,2.45,2.50,2.55},
  xticklabel style={/pgf/number format/fixed, /pgf/number format/precision=1},
  yticklabel style={/pgf/number format/fixed, /pgf/number format/precision=2},
  xlabel={Diversity ($1-$sim)}, xlabel style={yshift=3pt},
  ylabel={Judge overall}, ylabel style={yshift=-4pt},
  xmajorgrids,
  legend columns=4, legend style={at={(0.5,1.03)}, anchor=south, font=\small, /tikz/every even column/.append style={column sep=5pt}},
  legend image code/.code={\draw[#1] plot coordinates {(0.1cm,0.05cm)};},
]
\addplot[only marks, mark=square*, mark size=4.8pt, mark options={fill=nonefill, draw=nonedark, line width=0.4pt}] coordinates {(0.068,2.42) (0.048,2.45)};
\addlegendentry{None}
\addplot[only marks, mark=triangle*, mark size=5.7pt, mark options={fill=inferfill, draw=inferdark, line width=0.4pt}] coordinates {(0.116,2.46) (0.128,2.48)};
\addlegendentry{Foresight}
\addplot[only marks, mark=diamond*, mark size=5.7pt, mark options={fill=bothfill, draw=bothdark, line width=0.4pt}] coordinates {(0.158,2.51)};
\addlegendentry{Both}
\addplot[only marks, mark=*, mark size=4.8pt, mark options={fill=targetfill, draw=targetdark, line width=0.4pt}] coordinates {(0.215,2.50) (0.222,2.53)};
\addlegendentry{Hindsight}
\node[font=\footnotesize, nonedark, anchor=west, xshift=7pt] at (axis cs:0.068,2.42) {zero-shot};
\node[font=\footnotesize, nonedark, anchor=south, yshift=7pt] at (axis cs:0.048,2.45) {3-shot};
\node[font=\footnotesize, inferdark, anchor=north, yshift=-7pt] at (axis cs:0.116,2.46) {25K};
\node[font=\footnotesize, inferdark, anchor=west, xshift=7pt] at (axis cs:0.128,2.48) {50K};
\node[font=\footnotesize, bothdark, anchor=south, yshift=7pt] at (axis cs:0.158,2.51) {100K};
\node[font=\footnotesize, targetdark, anchor=west, xshift=7pt] at (axis cs:0.215,2.50) {25K};
\node[font=\footnotesize, targetdark, anchor=west, xshift=7pt] at (axis cs:0.222,2.53) {50K};
\draw[->, oc-gray-5, line width=0.5pt] (axis cs:0.02,2.535) -- (axis cs:0.06,2.55);
\node[font=\footnotesize, oc-gray-6, anchor=west] at (axis cs:0.06,2.55) {better};
\end{axis}
\end{tikzpicture}
\caption{Diversity vs.\ judge quality for \modelName-32B variants (labels give the number of training trajectories; \None is evaluated zero- and 3-shot). \Target is best on both axes, and more \Target data moves further up and to the right.}
\label{fig:ablation_scatter}
\end{figure}

\textbf{Metrics.}
We use four diversity metrics (Self-BLEU, embedding similarity, chunked BERTScore, and Sentence Mover's Similarity); definitions and full results are in Appendix~\ref{app:diversity}. Figures in the main text report embedding-based diversity, $1-$ the mean pairwise cosine similarity among the five trajectories generated for the same citation set.

\textbf{Results.}
The expectation is not borne out. \Target models achieve the highest diversity under \textit{every} metric in \textit{every} one of the eight model families (Table~\ref{tab:grouped_diversity_metrics}). Moreover, Figure~\ref{fig:ablation_scatter} shows that \Target models are simultaneously the highest-quality and the most diverse, and that adding more \Target data improves both. In other words, the extra structure of outcome-anchored supervision buys quality \emph{and} diversity at once.

\subsection{RQ3: Using Plans for Full Research Generation}
\label{sec:downstream_eval}

\begin{figure}[t]
\centering
\begin{tikzpicture}
\begin{axis}[
  scifig,
  width=0.92\linewidth, height=5.1cm,
  ybar=0.8pt, bar width=5pt,
  ymin=0, ymax=1.18, ytick={0,0.2,0.4,0.6,0.8,1.0},
  yticklabel style={/pgf/number format/fixed, /pgf/number format/precision=1},
  ylabel={Artifact score (LLM judge)}, ylabel style={yshift=-4pt},
  symbolic x coords={Exec,Ground,Util},
  xtick=data,
  xticklabels={Executability, {Scientific\\grounding}, {Downstream\\utility}},
  xticklabel style={font=\footnotesize, anchor=north, yshift=-1pt, align=center},
  enlarge x limits=0.14,
  nodes near coords, nodes near coords style={font=\scriptsize, /pgf/number format/fixed, /pgf/number format/fixed zerofill, /pgf/number format/precision=2, rotate=90, anchor=west, xshift=-1pt, color=oc-gray-8},
  legend columns=3, legend style={at={(0.5,1.02)}, anchor=south, column sep=3pt, font=\footnotesize},
]
\addplot[bas] coordinates {(Exec,0.61) (Ground,0.18) (Util,0.44)};
\addplot[fill=oc-gray-3, draw=basedark, line width=0.3pt, postaction={pattern=north east lines, pattern color=basedark}] coordinates {(Exec,0.73) (Ground,0.41) (Util,0.56)};
\addplot[non] coordinates {(Exec,0.79) (Ground,0.63) (Util,0.61)};
\addplot[inf] coordinates {(Exec,0.86) (Ground,0.70) (Util,0.67)};
\addplot[bth] coordinates {(Exec,0.88) (Ground,0.74) (Util,0.71)};
\addplot[tgt] coordinates {(Exec,0.91) (Ground,0.76) (Util,0.73)};
\addplot[cei] coordinates {(Exec,0.97) (Ground,0.89) (Util,0.93)};
\legend{No-Plan, Direct-Prompt, None, Foresight, Both, Hindsight, Teacher (ceiling)}
\end{axis}
\end{tikzpicture}
\caption{Downstream artifact quality when a coding agent is conditioned on each planning input (mean over five cited-work sets; LLM judge, 1--5 rescaled to $[0,1]$). Gray bars are prompting baselines without a distilled plan (dotted: No-Plan; hatched: Direct-Prompt), colored bars are \modelName-32B trajectories, and the dark bar is the teacher-model ceiling. Every distilled variant beats both baselines on all three criteria, and \Target is the best of the four conditions on each.}
\label{fig:downstream_eavl}
\end{figure}

To measure downstream impact, we hand each method's trajectory to a coding agent (\texttt{gpt-5.3-codex high}) that must produce a runnable repository and a short workshop-style paper from the same cited-work context. We compare the four \modelName-32B conditions (\None, \Infer, \Both, \Target) against two prompting baselines---\textbf{No-Plan} (cited papers only) and \textbf{Direct-Prompt} (cited papers plus a human-written planning prompt)---and a \textbf{Teacher} upper bound (trajectories from the 235B teacher). Artifacts are scored by an LLM judge from a different model family on three aggregate criteria, averaged over five cited-work sets: \emph{Executability}, \emph{Scientific Grounding}, and \emph{Downstream Utility} (Appendix~\ref{app:downstream_eval}). Figure~\ref{fig:downstream_eavl} (summarized in Figure~\ref{fig:overview_c}) shows that every distilled variant outperforms both baselines on every criterion, with the largest gains in scientific grounding and consistent gains in executability and utility. \Target is again the strongest condition.

\subsection{Further Analysis of Sensemaking-based Research Trajectories}

\begin{figure}[t]
\centering
\begin{tikzpicture}
\begin{axis}[
  scifig,
  width=0.95\linewidth, height=5.9cm,
  xbar=0pt, bar width=4.6pt,
  xmin=0, xmax=26, xtick={0,5,10,15,20,25},
  xticklabel={\pgfmathprintnumber{\tick}\%},
  xlabel={Share of trajectory annotations}, xlabel style={yshift=3pt},
  xmajorgrids, ymajorgrids=false,
  symbolic y coords={Presentation,Reframe,Questioning,Elaboration,Hypothesis,Schema,Shoebox,Foraging},
  ytick=data,
  yticklabels={
    Presentation,
    Reframe,
    Questioning,
    Elaboration,
    Hypothesis,
    Schema,
    Shoebox,
    Foraging},
  yticklabel style={font=\footnotesize, xshift=1pt},
  enlarge y limits=0.08,
  y axis line style={draw=none}, ytick style={draw=none},
  legend columns=2, legend style={at={(0.5,1.02)}, anchor=south, column sep=3pt, font=\footnotesize},
  nodes near coords, nodes near coords style={font=\scriptsize, /pgf/number format/fixed, /pgf/number format/fixed zerofill, /pgf/number format/precision=1, xshift=-1pt, color=oc-gray-7},
]
\addplot[inf] coordinates {(20.7,Presentation) (6.8,Reframe) (9.5,Questioning) (15.0,Elaboration) (13.6,Hypothesis) (17.2,Schema) (8.9,Shoebox) (8.3,Foraging)};
\addplot[tgt] coordinates {(18.6,Presentation) (5.2,Reframe) (9.3,Questioning) (16.7,Elaboration) (9.2,Hypothesis) (21.4,Schema) (7.6,Shoebox) (12.0,Foraging)};
\legend{Foresight, Hindsight}
\end{axis}
\end{tikzpicture}
\caption{Share of trajectory content annotated with each sensemaking stage (1K trajectories per condition, \modelName-32B). \Target spends more on the constructive stages (Schema, Foraging, Elaboration); \Infer on Hypothesis and Presentation.}
\label{fig:sensemaking_arcs}
\end{figure}

\textbf{Sensemaking Stage Distributions.}
Why do the two conditions differ? We annotate each segment of a trajectory with its sensemaking stage (Table~\ref{main_tab:sensemaking_stages}) and measure how much of the trajectory each stage occupies, over 1K generations each from \modelName-32B \Target and \Infer (Figure~\ref{fig:sensemaking_arcs}). \Target plans concentrate on the constructive middle stages: \textit{Schema} ($+4.2$ points over \Infer) and \textit{Foraging} ($+3.7$), consistent with a reconstruction objective that rewards building and filling coherent representational structures; \textit{Elaboration} also favours \Target ($+1.7$). \Infer plans allocate more to \textit{Hypothesis} ($+4.4$) and \textit{Presentation} ($+2.1$), reflecting open-ended ideation in which the model must generate and interrogate claims without a known target.

\textbf{Case Study.}
Figure~\ref{fig:framework-comparison} contrasts \Target and \Infer trajectories generated from the same face-attribute citation neighbourhood (stage labels follow Table~\ref{main_tab:sensemaking_stages}). The \Target trajectory is schema-first and artifact-grounded: it fixes datasets, constraints, and implementation details early. The \Infer trajectory opens with a high-level hypothesis and explores a broader, more speculative design space before settling on a concrete formulation. Further paired and unpaired excerpts are shown in Figure~\ref{fig:cached-example-sensemaking-panels} (Appendix~\ref{app:excerpts}).

\begin{figure*}[t]
\centering

\noindent
\begin{subfigure}[b]{0.48\linewidth}
\footnotesize
\vspace{1pt}
\begin{mdframed}[
  linecolor=targetdark!55, linewidth=1.2pt, backgroundcolor=targetfill!10,
  innerleftmargin=6pt, innerrightmargin=6pt,
  innertopmargin=5pt, innerbottommargin=5pt,
  skipabove=1pt, skipbelow=2pt]
\noindent\textbf{\S01 — Problem Definition \& Scope}
\vspace{1pt}\noindent
\scalebox{0.7}{\Tforag}
\textit{Rigorously define face attribute prediction under strict computational constraints, targeting real-world deployability.}
Key criteria: quantifiable efficiency metrics, prioritised over broader facial-analysis tasks.
\vspace{1pt}
\marginanno{\scalebox{0.7}{\Tshoebox}}{\#01}{%
  \textit{Define attribute set and source dataset as CelebA
  (40 attributes: gender, age, smile, glasses, beard, bald).}
  Ensures comparability with the 91.24\% accuracy claim.}
\vspace{1pt}\noindent
\scalebox{0.7}{\Tschema}
\textit{Quantify constraints: ${<}0.5$M params
(25$\times$ reduction), FLOPs ${<}50$M, memory ${>}87\%$ reduction.}
Concrete, falsifiable goals from the paper's claims.
\vspace{1pt}
\marginanno{\scalebox{0.7}{\Tschema}}{\#03}{%
  \textit{Slim Module:
  Input $\to$ DW Conv $(k{\times}k)$ $\to$ BN $\to$ ReLU
  $\to$ PW Conv $(1{\times}1)$ $\to$ BN $\to$ ReLU $\to$ Output.}
  Mirrors the paper's description.}
\vspace{1pt}\noindent
\scalebox{0.7}{\Telab}
\textit{Knowledge distillation: train ResNet-50 teacher on CelebA,
then train Slim-Net student with cross-entropy + KL-divergence loss.}
Transfers knowledge from powerful but inefficient model.
\end{mdframed}
\subcaption{\Target (anchored to the target paper)}\label{fig:framework-comparison-a}
\end{subfigure}
\hfill
\begin{subfigure}[b]{0.48\linewidth}
\footnotesize
\vspace{1pt}
\begin{mdframed}[
  linecolor=inferdark!55, linewidth=2.4pt, topline=false, rightline=false, bottomline=false, backgroundcolor=inferfill!10,
  innerleftmargin=6pt, innerrightmargin=6pt,
  innertopmargin=5pt, innerbottommargin=5pt,
  skipabove=1pt, skipbelow=2pt]
\noindent\textbf{\S01 — Hypothesised Breakthrough}
\vspace{1pt}\noindent
\scalebox{0.7}{\Thyp}
\textit{A unified multi-task deep learning framework for facial
attribute recognition in unconstrained environments,
addressing attribute heterogeneity and label imbalance.}
\vspace{1pt}
\marginanno{\scalebox{0.7}{\Tschema}}{\#01}{%
  \textit{Novelty: mixed objective optimisation with adaptive
  loss re-weighting for joint learning without compromising
  rare-label performance.}}
\vspace{1pt}\noindent
\scalebox{0.7}{\Telab}
\textit{Architecture builds on a deep CNN backbone ---
Inception-style or ResNeXt --- with shared and
task-specific feature pathways.}
\vspace{1pt}\noindent
\scalebox{0.7}{\Tforag}
\textit{Analyse facial attribute types by semantic scope
(local vs.\ global) and distribution (balanced vs.\ imbalanced)
to define task groups and design a modular network.}
\vspace{1pt}
\marginanno{\scalebox{0.7}{\Tquest}}{\#04}{%
  \textit{Open Question: How to optimally partition the network for
  maximal cross-task transfer without negative interference?}}
\vspace{1pt}\noindent
\scalebox{0.7}{\Tschema}
\textit{CNN backbone with Inception or ResNeXt blocks.
Mixed objective loss with domain-adaptive re-weighting.}
\end{mdframed}
\subcaption{\Infer (open-ended, no anchor)}\label{fig:framework-comparison-b}
\end{subfigure}

\caption{Excerpts of (a)~\Target and (b)~\Infer trajectories generated from the same citation neighbourhood. \Target excerpts use a full frame and \Infer excerpts a left bar only. Pill labels mark sensemaking stages; gray boxes are numbered side annotations.}
\label{fig:framework-comparison}
\end{figure*}

\section{Related Work}

Table~\ref{tab:framework_comparison} situates \frameworkName among existing scientific research agents along four dimensions.

\begin{table}[t]
\centering
\small
\setlength{\tabcolsep}{3pt}
\renewcommand{\arraystretch}{1.15}
\providecommand{\hdr}[2]{\begin{tabular}[b]{@{}c@{}}#1\\#2\end{tabular}}
\begin{tabular}{@{}lcccc@{}}
\toprule
\textbf{Framework} & \hdr{Sense-}{making} & \hdr{Citation-}{cond.} & \hdr{Scalable}{data} & \hdr{Open}{source} \\
\midrule
AutoScience~\citeyearpar{autoscience2025carl} & \xmark & \xmark & \pmark & \xmark \\
AI Scientist~\citeyearpar{lu2024ai} & \xmark & \pmark & \xmark & \cmark \\
ResearchAgent~\citeyearpar{baek2025researchagent} & \pmark & \pmark & \xmark & \cmark \\
SciAgent~\citeyearpar{ma2024sciagenttoolaugmentedlanguagemodels} & \pmark & \xmark & \xmark & \cmark \\
\midrule
\rowcolor{oc-gray-1}
\textbf{\frameworkName (Ours)} & \cmark & \cmark & \cmark & \cmark \\
\bottomrule
\end{tabular}
\caption{Comparison with existing scientific research frameworks. \emph{Sense-making}: ideation follows an explicit multi-stage cognitive scaffold; \emph{citation-cond.}: ideation is grounded only in a paper's cited works; \emph{scalable data}: releases a large trajectory corpus for training. \cmark~full, \pmark~partial, \xmark~none.}
\label{tab:framework_comparison}
\end{table}

\begin{comment}
    
Our work sits at the intersection of agentic research systems, procedural sensemaking over literature, and knowledge distillation for scientific reasoning. Unlike prior work that treats research as an end-to-end generation task, we isolate the planning stage and ask: can we \emph{learn and evaluate the research procedure itself} when supervision comes solely from a paper's citation neighborhood? This framing---citation-conditioned trajectory generation---provides a controlled lens on scientific creativity that is absent from existing systems.

\end{comment}

\textbf{Scientific Research Agents}
Web-enabled ``deep research'' agents---Tongyi DeepResearch~\citep{tongyi_deepresearch_2025}, DR Tulu~\citep{dr_tulu_2025}, and \textsc{TTD-DR} \citep{ttd_dr_2025}---excel at end-to-end report generation but operate as black boxes, making it unclear which part of the research loop improves and why. Towards autonomous scientific innovation, \textsc{AI-Researcher} \citep{ai_researcher_2025} and Google's \textsc{AI co-scientist} \citep{ai_coscientist_arxiv_2025} target hypothesis generation and literature-to-writing pipelines, while \citet{barbarians_gate_2025} argue that closed-loop optimization requires strong verifiers. Evolutionary-search approaches instead learn to discover by iteratively refining candidate solutions, either by fine-tuning on large collections of search trajectories \citep{lee2026evolution} or by co-evolving web search with task solving \citep{lee2026evoduet}. Our work is complementary: rather than automating the full pipeline, we specifically operationalize the \emph{ideation} stage through structured, citation-grounded trajectories, providing a controlled supervision signal these systems lack.

\textbf{Process Data for Expert Work}
A complementary line of work records the \emph{process} behind expert artifacts rather than only the artifacts themselves: keystroke-level trajectories of scholarly writing \citep{koo2023decoding,le2026scholawrite}, lawyers' step-by-step reasoning on case studies \citep{das2025lawflow}, and creators' workflows for communicating science \citep{park2025stealing}. These studies find that expert processes are structured and iterative, whereas LLMs tend toward shallow, single-pass outputs---for example, synthesizing cited work only superficially when composing related-work sections \citep{martinboyle2024shallow}. \frameworkName brings this process view to research ideation by distilling multi-stage sensemaking trajectories conditioned on a paper's citations.

\textbf{Sensemaking and Research Evaluation}
Existing benchmarks measure research abilities in realistic settings---\textsc{LiveResearchBench} \citep{liveresearchbench_2025} targets coverage and citation quality, \textsc{DeepScholar-Bench} \citep{deepscholarbench_2025} emphasizes nugget coverage and attribution, and \textsc{AStaBench} \citep{astabench_2025} spans literature understanding through end-to-end workflows---but typically entangle multiple skills, making it hard to isolate procedural planning. Our framework factors sensemaking into a single controllable primitive: given only cited works, generate a trajectory that either reconstructs how citations led to the target paper or proposes grounded forward directions, enabling clean comparisons across supervision regimes. Work on MLE-bench trajectories \citep{mle_bench_2024} finds that broader initial exploration correlates with stronger downstream outcomes---a finding we both confirm and complicate, showing that \emph{structured} reconstruction produces more diverse outputs than open-ended ideation. Evaluating generated research is itself difficult: AI-written papers often contain cross-section inconsistencies that surface-level detectors miss \citep{oh2026scienceslop}, so we score both the trajectories and the artifacts built from them.

\section{Conclusion}
We introduced citation-conditioned research trajectories as a controllable primitive for studying scientific ideation in LLMs, isolating the sensemaking stage that precedes implementation and enabling direct comparison between open-ended exploration and structured reconstruction.
Our central finding is counterintuitive: models trained on constrained, outcome-anchored \Target trajectories produce outputs that are \emph{more} diverse and \emph{more} novel than those trained explicitly for open-ended exploration (\Infer)—an advantage that propagates downstream into stronger execution success, tighter plan–artifact alignment, and higher judged quality.

We attribute this to the structure of the training signal itself. Each \Target example is anchored to a distinct real paper whose contribution lies at some displacement from its citation neighbourhood, so the student internalises a distribution over \emph{diverse attractor points} and generates trajectories that move toward distinctive contributions. \Infer training, by contrast, anchors every example to a single teacher model's generative prior, so distillation amplifies the teacher's structural regularities rather than the heterogeneity of real research, consistent with evidence that LLM-generated data carries systematic artifacts of its generator \citep{das2024surface}.

\section*{Limitations}
Our work is limited in scope in several ways. First, the downstream evaluation covers only a small sample of computer science tasks; broader coverage across domains and problem types is needed to establish robustness. Second, our distillation results are restricted to supervised fine-tuning; reinforcement learning (RL) from human or AI feedback on trajectory quality may yield further improvements (although we apply small RL experiments), they are rather limited in scope. Third, the dataset and models used for generating the reasoning trajectories were fixed to only one model (Qwen235-A22b) and dataset (s2orc). Further evaluations should explore a wider set of models and a larger set of inputs (for example, incorporating other information beyond cited works such as github, blogs, or other datasets as well).

\section*{Ethical Considerations}
This work uses publicly available metadata and abstracts from the Semantic Scholar Open Research Corpus (S2ORC)~\citep{s2orc}, which is released under the Semantic Scholar Dataset License Agreement. No private, proprietary, or personally identifiable data was collected or used. All trajectory generation was performed using publicly available LLMs, and no human subjects were involved beyond the voluntary participation of graduate-student co-authors in the annotation study described in \cref{sec:quality}.

Generating 120K sensemaking trajectories with a 235B-parameter teacher model, followed by fine-tuning eight model families and running RL post-training, incurs non-trivial computational cost and associated energy consumption. We mitigate this by training each SFT model for only one epoch and restricting RL post-training to the two strongest checkpoints.

We release \datasetName, \modelName weights, and all prompt templates to support reproducibility and downstream research. While sensemaking-based trajectories are designed to improve the quality of AI-assisted scientific ideation, we acknowledge that such tools could be misused to generate plausible-sounding but unverified research proposals at scale. We encourage users to treat generated trajectories as starting points for human-guided research rather than as finished scientific contributions.

\ifarxiv
\section*{Acknowledgments}
This material is based upon work supported by the National Science Foundation (NSF) and the National Artificial Intelligence Research Resource (NAIRR) under Grant No. NAIRR240308. Any opinions, findings, and conclusions or recommendations expressed in this material are those of the authors and do not necessarily reflect the views of NSF or NAIRR.
\fi

\bibliography{colm2026_conference}

\appendix
\newpage
\newpage

\section{Citation Neighborhood Pipeline Details}
\label{app:pipeline}

We illustrate the full multi-stage pipeline:
1) extracting citation neighborhoods from S2ORC;
2) summarizing individual papers;
3) generating \Target and \Infer research trajectories;
4) enforcing structured outputs via prompting constraints.

\subsection{YAML prompt template libraries}
We package our prompt text as YAML template libraries to keep the pipeline reproducible and easy to audit: prompt wording is versioned separately from orchestration code, and each stage calls a named template with explicit inputs. Listing~\ref{lst:analysis-yaml} (\texttt{analysis\_prompts.yaml}) contains templates for structured paper- and citation-network analysis (e.g., role-based paper summaries, cluster profiling, and gap finding). Listing~\ref{lst:insight-yaml} (\texttt{insight\_prompts.yaml}) contains templates for cross-cluster synthesis, prompting the model to generate “cross-pollination” research ideas by connecting complementary clusters and transferring methods across areas.

\begin{lstlisting}[
  style=paper-code,
  caption={YAML prompt set for paper- and citation-network analysis (e.g., cluster profiling, gap finding).},
  label={lst:analysis-yaml}
]
paper_analysis:
  foundational:
    template: |
      You are reviewing a foundational paper in its field.
      Title: {title}
      Abstract: {abstract}
      Known citations: {citing_papers}
      Citation count: {citation_count}

      Provide a concise expert summary covering:
      - Why this work shaped the field
      - Which assumptions or design choices enabled its impact
      - What elements no longer hold
  bridge:
    template: |
      Analyze this bridge paper that connects disparate subfields.
      Title: {title}
      Abstract: {abstract}
      Influenced papers: {citing_papers}

      Describe:
      - The conceptual translation it performs
      - How it reframes terminology or methods
      - What tensions remain unresolved
  frontier:
    template: |
      Summarize this frontier paper.
      Title: {title}
      Abstract: {abstract}
      Early citations: {citing_papers}

      Explain:
      - The emerging problem it addresses
      - How it builds on recent momentum
      - Where validation is still thin
  method:
    template: |
      Examine this methodological paper.
      Title: {title}
      Abstract: {abstract}
      Downstream users: {citing_papers}

      Detail:
      - The core methodological advance
      - Typical contexts where it excels
      - Failure modes or assumptions to test
  supporting:
    template: |
      Provide a brief expert summary of the following paper.
      Title: {title}
      Abstract: {abstract}

      Focus on:
      - Central claim
      - Evidence cited
      - Key dependencies

paper_contributions:
  foundational:
    template: |
      Distill the lasting contributions of this foundational work.
      Title: {title}
      Built upon: {building_papers}
      Builds on: {backbone_papers}

      Highlight:
      - Breakthrough idea
      - Enduring techniques
      - Concepts others adapted
  bridge:
    template: |
      Identify the synthesis that this bridge paper enabled.
      Title: {title}
      Builds on: {backbone_papers}
      Follow-on work: {building_papers}

      Capture:
      - Cross-domain translation
      - Tooling or frameworks introduced
      - How it changed the direction of citing papers
  frontier:
    template: |
      Extract the contributions of this frontier paper.
      Title: {title}
      Builds on: {backbone_papers}

      Describe:
      - Novel problem framing
      - Prototype methods or datasets
      - Hypotheses pending validation
  method:
    template: |
      Summarize the methodological toolkit from this paper.
      Title: {title}
      Builds on: {backbone_papers}
      Downstream use: {building_papers}

      Include:
      - Core algorithm or protocol
      - Required inputs and outputs
      - Extensions that citing papers added
  supporting:
    template: |
      Note the supporting contributions of this paper.
      Title: {title}
      Builds on: {backbone_papers}

      Address:
      - Specific niche it serves
      - How it reinforces accepted results
      - Any incremental advances

paper_gaps:
  foundational:
    template: |
      Examine blind spots of this foundational paper.
      Title: {title}
      Abstract: {abstract}

      Identify:
      - Assumptions that deserve re-testing
      - Datasets or contexts it skipped
      - Technical bottlenecks that still persist
  bridge:
    template: |
      Surface the limitations of this bridge paper.
      Title: {title}
      Abstract: {abstract}

      Cover:
      - Concepts that were mistranslated
      - Areas the bridge failed to integrate
      - Experiments or evaluations still missing
  frontier:
    template: |
      Flag the uncertainties within this frontier paper.
      Title: {title}
      Abstract: {abstract}

      Discuss:
      - Uncontrolled variables
      - Dependencies on unstable baselines
      - Steps needed for broad adoption
  method:
    template: |
      Analyze the weak spots of this methodological work.
      Title: {title}
      Abstract: {abstract}

      Outline:
      - Scalability constraints
      - Sensitivity to hyperparameters or data
      - Generalization risks
  supporting:
    template: |
      Note gaps or follow-up questions raised.
      Title: {title}
      Abstract: {abstract}

      Include:
      - Underexplored use cases
      - Missing comparisons
      - Simplifying assumptions that may fail

network_insights:
  cluster_profile:
    template: |
      Assess research cluster {cluster_id}.
      Representative papers: {paper_titles}
      Key exemplars: {representative_papers}
      Timeline: {timeline}

      Provide:
      - Dominant questions
      - Shared methodologies
      - Internal disagreements
      - Signals of maturity or stagnation
  gap_analysis:
    template: |
      Detect research gaps for cluster {cluster_id}.
      Core papers: {paper_titles}

      Report:
      - Methods not yet cross-applied
      - Contradictory findings to reconcile
      - High-value datasets or benchmarks still missing
\end{lstlisting}

\begin{lstlisting}[
  style=paper-code,
  caption={YAML prompt set for cross-cluster synthesis and generating “cross-pollination” research ideas.},
  label={lst:insight-yaml}
]
cross_pollination:
  novel_connections:
    template: |
      You have a citation network with clusters grouped by role:
      {major_clusters}

      Temporal activity profile:
      {temporal_flow}

      Produce novel research directions by:
      - Linking complementary clusters
      - Transferring mature methods to rising areas
      - Challenging dominant assumptions

      Return bullets that can inspire concrete projects.
\end{lstlisting}

\subsection{Prompt Structures}
We use controlled prompt templates to elicit consistent, comparable outputs from the teacher model across ideation modes. Listing~\ref{lst:prompt-inferred} shows the \Infer prompt: given only cited-paper synopses (with the target text withheld), the model must hypothesize a plausible breakthrough and produce a staged research program, grounding each inference with inline paper-ID citations. Listing~\ref{lst:prompt-target} shows the complementary \Target (reconstruction) prompt: given a prescribed section outline and target synopsis, the model generates an executable trajectory with concrete checkpoints paired with meta-level rationales, enabling a faithful reconstruction-style trajectory that remains auditable and systematically comparable to the \Infer setting.

\begin{lstlisting}[
  style=paper-code,
  caption={Foresight prompt: generate a hypothesized breakthrough + research program from cited synopses only.},
  label={lst:prompt-inferred}
]
You are an AI research strategist tasked with inferring the most probable and high-impact research plan based solely on the cited works provided (the target paper text is unavailable). Produce a two-part Markdown document that hypothesizes what the target contribution could be and charts a credible path to deliver it.

Output format:
1. A section titled ``Hypothesized Breakthrough'' (rendered as `## Hypothesized Breakthrough`) that summarizes:
   - The likely research objective and novelty inferred from the cited corpus.
   - The expected system / method architecture.
   - The anticipated evaluation setting and success criteria.
   Include inline references to the relevant cited papers using [[paper_id]] to justify each inference.
2. A section titled ``High-Impact Research Program'' (rendered as `## High-Impact Research Program`) with subsections for Idea Formulation, Modeling, Experimentation, Validation, and Analysis (render each as `###` headings). For each subsection:
   - Articulate specific actions, checkpoints, and decision gates that synthesize insights from the cited works.
   - Include meta-level rationales that explain why the step is essential or risky, citing the supporting papers with [[paper_id]].
   - Highlight any contingency plans or open questions implied by the cited works.

Citation rule: every time you lean on a cited paper for evidence, design inspirations, cautions, or evaluation criteria, cite it inline using [[paper_id]].

Example snippet (do not reuse verbatim):
- Hypothesized Objective: Unify latent dense retrieval with multi-hop reasoning to handle knowledge gaps in open QA [[173990818]], [[52822214]].

Cited paper synopses:
{cited_synopses}

Produce the requested Markdown now.
\end{lstlisting}

\begin{lstlisting}[
  style=paper-code,
  caption={Hindsight (reconstruction) prompt: turn the target outline/synopsis into an executable trajectory with checkpoints + meta rationales.},
  label={lst:prompt-target}
]
You are an AI research strategist tasked with turning the target paper into a reproducible, high-agency research program. Follow the section outline supplied below and write a single Markdown document that blends execution guidance with meta-level reasoning.

Section outline (use these headings verbatim, in order, expanding each with the requested depth):
{structure_plan}

For every section and subsection:
- Begin with a short paragraph that states the intent of that stage and the key decision criteria, and briefly explain why these criteria were prioritized instead of obvious alternatives. Cite the motivating papers using [[paper_id]].
- Follow with 1-3 bullet points detailing concrete actions or checkpoints (the implementation blueprint), each paired with a ``Meta rationale''' sentence that exposes the authors' or your inferred reasoning (e.g., trade-offs, precedent, or risk mitigation) and cites the relevant papers.
- Where a decision depends directly on the target paper, describe the rationale in prose (e.g., ``the canonical pipeline prioritizes retrieval transparency'') and contrast it with at least one cited work using [[paper_id]].
- Surface assumptions, dependencies, and fallback options for the step; note any conditions that would cause you to revisit the decision.

Global requirements:
- Interpret each top-level bullet from the outline as an H2 heading (`## {{title}}`) and each nested bullet as an H3 heading (`### {{title}}`). Do not reproduce the bullet list; use Markdown headings only.
- Keep the tone analytic and directive; make it easy for a research team to follow without losing the ``why.''
- Cite external papers using [[paper_id]]; do not emit [[target]] or other placeholders for the target work-reference it descriptively instead.
- Highlight potential parallelization or sequencing constraints when relevant (e.g., checkpoints that can run concurrently versus those gated by data availability).
- Never mention the target paper's title, authors, venue, or publication year; refer to it only implicitly.

Context for reference:
- Target synopsis (for reference only; do not restate metadata): {target_synopsis}
- Cited paper synopses: {cited_synopses}

Produce only the Markdown document that follows the prescribed section structure.
\end{lstlisting}

\section{Citation Statistics}
\label{app:cite-stat}

A central requirement of citation-conditioned research is that models must reason over prior work in a disciplined and appropriate manner. We therefore evaluate citation behavior along two axes: \emph{structural similarity} to the teacher trajectory and \emph{semantic appropriateness} of citation usage.

\paragraph{Citation Similarity Metrics}
We first measure how closely the student trajectory matches the citation structure of the teacher trajectory. For each generated trajectory, we parse cited paper identifiers and compute:
(i) the L$_1$ absolute difference in total citation count between $P^{(T)}$ and $P^{(S)}$;
(ii) the L$_1$ difference in the number of unique cited papers;
(iii) the Jaccard distance between the two citation sets:
\begin{align*}
&\text{JaccardDistance}(C^{(T)},C^{(S)}) \\
&\qquad = 1 - \frac{|C^{(T)} \cap C^{(S)}|}{|C^{(T)} \cup C^{(S)}|}.
\end{align*}
These metrics quantify whether student models attend to a similar body of evidence as the teacher when constructing research trajectories.

\begin{table*}[ht]
\centering
\scriptsize
\setlength{\tabcolsep}{2pt}
\resizebox{\textwidth}{!}{%
\begin{tabular}{lrrrrrrcccc}
\toprule
\textbf{Checkpoint} & \textbf{Len$_{\text{char}}$ (Fore)} & \textbf{Len$_{\text{char}}$ (Hind)} & \textbf{Cites (Fore)} & \textbf{Cites (Hind)} & \textbf{Uniq (Fore)} & \textbf{Uniq (Hind)} & \textbf{Rep$_2$ (Fore)} & \textbf{Rep$_2$ (Hind)} & \textbf{Rep$_{16}$ (Fore)} & \textbf{Rep$_{16}$ (Hind)} \\
\midrule
Q32-Hindsight & \heat{13919.258}{11016.898}{17499.652} & \heat{14027.972}{11760.246}{16914.440} & \heat{29.086}{29.086}{55.633} & \heat{28.082}{28.082}{55.485} & \heat{5.906}{5.906}{9.641} & \heat{5.515}{5.515}{9.082} & \heat{0.240}{0.188}{0.350} & \heat{0.247}{0.202}{0.348} & \heat{0.039}{0.007}{0.132} & \heat{0.041}{0.012}{0.107} \\
Q32-Foresight  & \heat{11016.898}{11016.898}{17499.652} & \heat{11760.246}{11760.246}{16914.440} & \heat{36.180}{29.086}{55.633} & \heat{35.187}{28.082}{55.485} & \heat{5.977}{5.906}{9.641} & \heat{5.970}{5.515}{9.082} & \heat{0.188}{0.188}{0.350} & \heat{0.202}{0.202}{0.348} & \heat{0.007}{0.007}{0.132} & \heat{0.012}{0.012}{0.107} \\
Q32-Both   & \heat{11303.008}{11016.898}{17499.652} & \heat{12319.485}{11760.246}{16914.440} & \heat{37.023}{29.086}{55.633} & \heat{32.037}{28.082}{55.485} & \heat{5.914}{5.906}{9.641} & \heat{5.522}{5.515}{9.082} & \heat{0.190}{0.188}{0.350} & \heat{0.234}{0.202}{0.348} & \heat{0.012}{0.007}{0.132} & \heat{0.027}{0.012}{0.107} \\
\midrule
Q4-Hindsight  & \heat{16723.875}{11016.898}{17499.652} & \heat{15216.052}{11760.246}{16914.440} & \heat{36.961}{29.086}{55.633} & \heat{34.022}{28.082}{55.485} & \heat{9.641}{5.906}{9.641} & \heat{9.082}{5.515}{9.082} & \heat{0.350}{0.188}{0.350} & \heat{0.348}{0.202}{0.348} & \heat{0.132}{0.007}{0.132} & \heat{0.107}{0.012}{0.107} \\
Q4-I-Foresight & \heat{12327.438}{11016.898}{17499.652} & \heat{16914.440}{11760.246}{16914.440} & \heat{52.703}{29.086}{55.633} & \heat{53.284}{28.082}{55.485} & \heat{5.977}{5.906}{9.641} & \heat{6.082}{5.515}{9.082} & \heat{0.242}{0.188}{0.350} & \heat{0.307}{0.202}{0.348} & \heat{0.060}{0.007}{0.132} & \heat{0.098}{0.012}{0.107} \\
Q4-I-Both  & \heat{14708.986}{11016.898}{17499.652} & \heat{16173.500}{11760.246}{16914.440} & \heat{51.672}{29.086}{55.633} & \heat{48.052}{28.082}{55.485} & \heat{6.031}{5.906}{9.641} & \heat{5.828}{5.515}{9.082} & \heat{0.281}{0.188}{0.350} & \heat{0.314}{0.202}{0.348} & \heat{0.075}{0.007}{0.132} & \heat{0.082}{0.012}{0.107} \\
\midrule
Q303-Hindsight & \heat{16464.223}{11016.898}{17499.652} & \heat{15427.216}{11760.246}{16914.440} & \heat{42.962}{29.086}{55.633} & \heat{39.649}{28.082}{55.485} & \heat{8.156}{5.906}{9.641} & \heat{7.761}{5.515}{9.082} & \heat{0.323}{0.188}{0.350} & \heat{0.312}{0.202}{0.348} & \heat{0.097}{0.007}{0.132} & \heat{0.080}{0.012}{0.107} \\
Q303-I-Foresight& \heat{12413.791}{11016.898}{17499.652} & \heat{16871.356}{11760.246}{16914.440} & \heat{55.633}{29.086}{55.633} & \heat{55.485}{28.082}{55.485} & \heat{6.000}{5.906}{9.641} & \heat{5.985}{5.515}{9.082} & \heat{0.250}{0.188}{0.350} & \heat{0.309}{0.202}{0.348} & \heat{0.059}{0.007}{0.132} & \heat{0.100}{0.012}{0.107} \\
Q303-I-Both & \heat{15064.660}{11016.898}{17499.652} & \heat{16095.015}{11760.246}{16914.440} & \heat{54.287}{29.086}{55.633} & \heat{50.694}{28.082}{55.485} & \heat{6.164}{5.906}{9.641} & \heat{5.940}{5.515}{9.082} & \heat{0.286}{0.188}{0.350} & \heat{0.318}{0.202}{0.348} & \heat{0.074}{0.007}{0.132} & \heat{0.083}{0.012}{0.107} \\
Q303-Foresight  & \heat{15419.668}{11016.898}{17499.652} & \heat{15658.179}{11760.246}{16914.440} & \heat{44.094}{29.086}{55.633} & \heat{43.037}{28.082}{55.485} & \heat{7.352}{5.906}{9.641} & \heat{7.343}{5.515}{9.082} & \heat{0.292}{0.188}{0.350} & \heat{0.297}{0.202}{0.348} & \heat{0.071}{0.007}{0.132} & \heat{0.068}{0.012}{0.107} \\
Q303-Both   & \heat{15562.858}{11016.898}{17499.652} & \heat{15575.903}{11760.246}{16914.440} & \heat{43.344}{29.086}{55.633} & \heat{41.918}{28.082}{55.485} & \heat{7.180}{5.906}{9.641} & \heat{7.045}{5.515}{9.082} & \heat{0.291}{0.188}{0.350} & \heat{0.300}{0.202}{0.348} & \heat{0.073}{0.007}{0.132} & \heat{0.070}{0.012}{0.107} \\
Q303-None   & \heat{15285.785}{11016.898}{17499.652} & \heat{15252.127}{11760.246}{16914.440} & \heat{37.438}{29.086}{55.633} & \heat{35.649}{28.082}{55.485} & \heat{6.875}{5.906}{9.641} & \heat{6.985}{5.515}{9.082} & \heat{0.271}{0.188}{0.350} & \heat{0.286}{0.202}{0.348} & \heat{0.056}{0.007}{0.132} & \heat{0.060}{0.012}{0.107} \\
\midrule
Q4-Foresight    & \heat{16608.129}{11016.898}{17499.652} & \heat{15974.329}{11760.246}{16914.440} & \heat{42.180}{29.086}{55.633} & \heat{39.724}{28.082}{55.485} & \heat{7.531}{5.906}{9.641} & \heat{7.321}{5.515}{9.082} & \heat{0.306}{0.188}{0.350} & \heat{0.293}{0.202}{0.348} & \heat{0.074}{0.007}{0.132} & \heat{0.066}{0.012}{0.107} \\
Q4-Both     & \heat{16721.926}{11016.898}{17499.652} & \heat{15922.202}{11760.246}{16914.440} & \heat{41.625}{29.086}{55.633} & \heat{40.157}{28.082}{55.485} & \heat{7.398}{5.906}{9.641} & \heat{7.269}{5.515}{9.082} & \heat{0.304}{0.188}{0.350} & \heat{0.296}{0.202}{0.348} & \heat{0.074}{0.007}{0.132} & \heat{0.069}{0.012}{0.107} \\
Q4-None     & \heat{16419.836}{11016.898}{17499.652} & \heat{15606.097}{11760.246}{16914.440} & \heat{37.305}{29.086}{55.633} & \heat{35.590}{28.082}{55.485} & \heat{7.055}{5.906}{9.641} & \heat{7.052}{5.515}{9.082} & \heat{0.288}{0.188}{0.350} & \heat{0.283}{0.202}{0.348} & \heat{0.062}{0.007}{0.132} & \heat{0.060}{0.012}{0.107} \\
\bottomrule
\end{tabular}
}%
\caption{Generation statistics for \Infer vs.\ \Target ideations. We report average generation length (chars), citation usage (total/unique), and repetition proportions (2-gram and 16-gram). Cell color indicates per-column magnitude (darker = larger).}
\label{tab:generation_statistics}
\end{table*}

Table~\ref{tab:generation_statistics} summarizes how our different SFT formatting/generation regimes trade off length, citation density, and repetition. Across checkpoints, the \texttt{I-Foresight} setting tends to allocate a larger fraction of the budget to citations (higher \# citations and comparable or slightly higher unique citations) while remaining relatively concise in total characters. In contrast, \texttt{Hindsight} generations are typically longer in raw character count but do not consistently increase citation coverage, suggesting that additional non-citation context dominates the extra budget. The repetition metrics (\textit{Rep}$_2$ and \textit{Rep}$_{16}$) generally increase with longer outputs, indicating that expanding the character budget can add redundant text rather than new evidence.

\section{SFT Ablations}
\label{app:sft}

\begin{figure}[ht]
  \centering
  \includegraphics[width=\columnwidth]{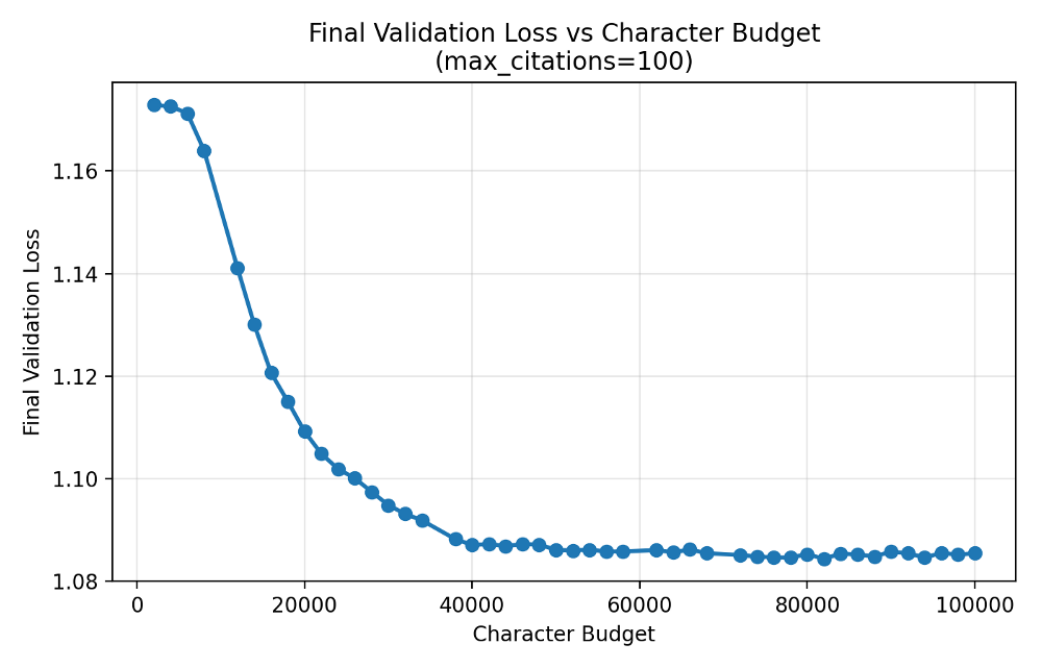}
  \caption{\textbf{SFT ablation: character budget (max\_citations=100).}
  Increasing the prepended character budget reduces validation loss initially, then saturates, indicating diminishing returns from additional non-citation context.}
  \label{fig:sft-charbudget}
\end{figure}

\begin{figure}[ht]
  \centering
  \includegraphics[width=\columnwidth]{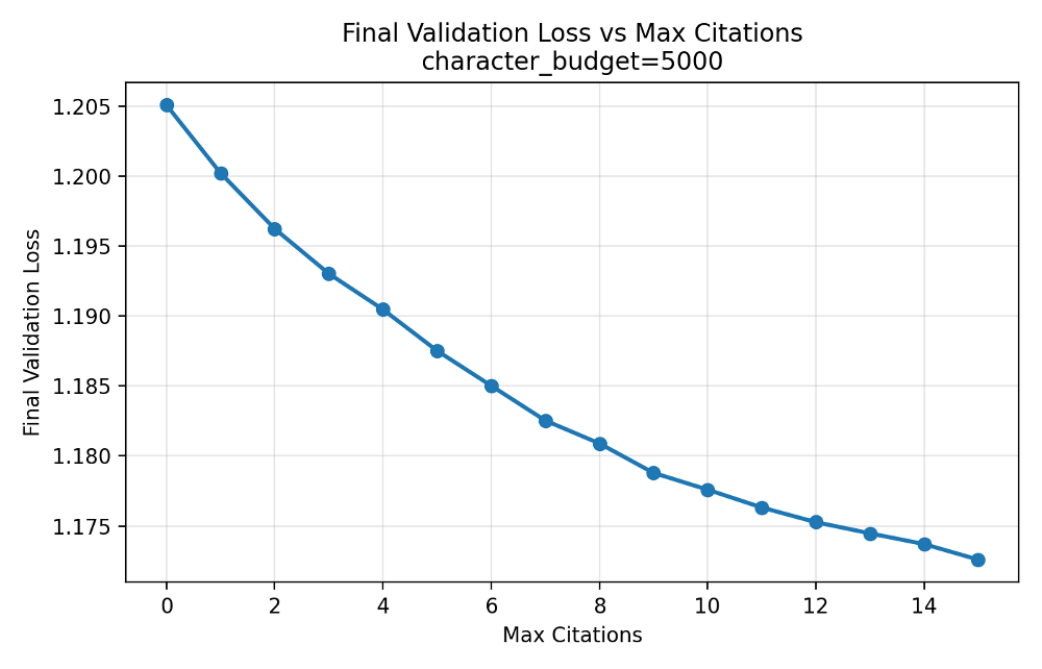}
  \caption{\textbf{SFT ablation: citation cap (character budget = 5k).}
  Under a fixed character budget, allowing more citations consistently improves validation loss, motivating our choice to maximize citation retention while pruning surrounding text.}
  \label{fig:sft-maxcites}
\end{figure}

We ablate two SFT formatting knobs: the amount of prepended context (character budget) and the maximum number of citations retained.
Figure~\ref{fig:sft-charbudget} shows that expanding the character budget helps early but quickly plateaus when citation capacity is already high.
In contrast, Figure~\ref{fig:sft-maxcites} shows that, under a fixed 5k character budget, increasing the citation cap continues to reduce validation loss.
Together, these trends justify allocating limited context to \emph{citations} and pruning non-citation characters.

\section{Evaluation Protocol Overview}
\label{app:eval_protocol}
\begin{figure*}[t]
\centering
\includegraphics[width=\textwidth]{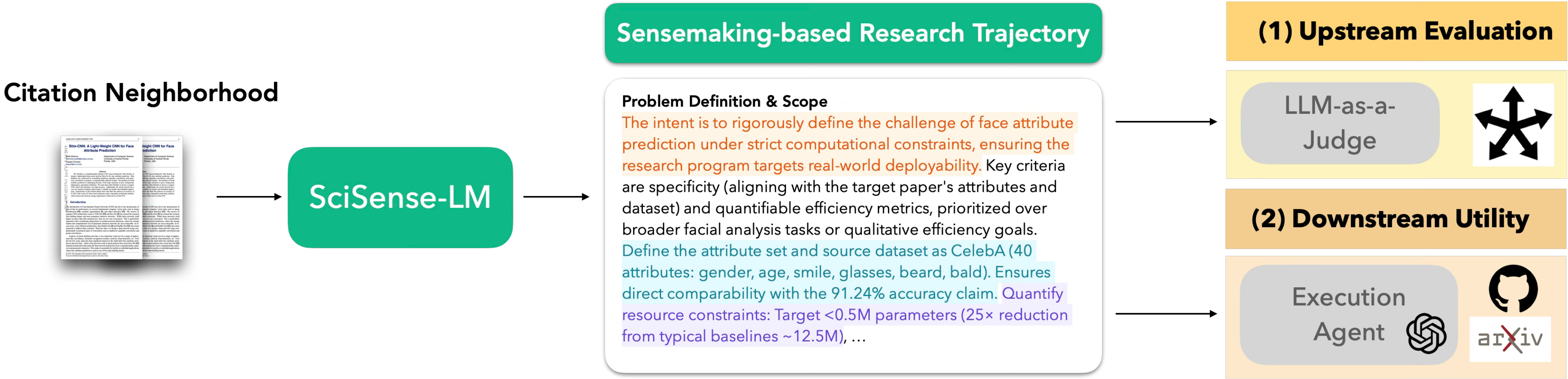}
\caption{Evaluation protocol. Given a citation neighbourhood, \modelName generates a sensemaking trajectory that is assessed (1)~\emph{upstream}, on its own quality (LLM judge) and diversity, and (2)~\emph{downstream}, by handing it to a coding agent and judging the resulting code and paper artifacts.}
\label{fig:eval_pipeline}
\end{figure*}

Figure~\ref{fig:eval_pipeline} summarizes the two evaluation setups used in Section~\ref{main_sec:experiments}: upstream evaluation of the trajectory itself (LLM-judge quality and sample diversity) and downstream evaluation of the artifacts a coding agent produces when conditioned on the trajectory.

\section{Downstream Research Generation Details}
\label{app:downstream_eval}
We describe the procedure used to generate outputs and evaluation results in Section \ref{sec:downstream_eval}.
\subsection{Downstream Research Artifact Generation}

\paragraph{Filtering and Selection of Research Plans.}
We began from a pool of automatically generated research plans, where each plan belonged to one of four supervision conditions: \None, \Infer, \Target, and \Both. For each underlying seed instance, we retained only cases for which all four conditions produced corresponding plans, enabling matched qualitative comparison across supervision regimes. We then manually filtered these matched sets using four criteria. First, the proposed research trajectory had to fall within computer science or otherwise be executable entirely through computation, as our evaluation setting does not support physical experimentation or human-subject actuation. Second, the plan had to appear feasible under modest compute constraints, excluding projects that would likely require prohibitively large-scale training or data collection. Third, the core artifacts required by the plan, including datasets, pretrained models, and software dependencies, had to be publicly available. These artifacts were downloaded in advance into a shared local repository so that downstream execution would not depend on open-ended discovery. Fourth, we required that all retained plan sets preserve correspondence across the previous three conditions for the same seed context. The resulting benchmark is therefore intentionally small and manually curated, prioritizing feasibility and controlled comparison over scale. This yielded a total of 5 unique citation sets, with 4 generated trajectories from our distill models from the \modelName-32B class.

\paragraph{Agentic Execution Environment.}
For each retained instance, we instantiated an autonomous coding workflow in which a coding agent was provided with: (i) an optional single research trajectory, (ii) the cited papers referenced by that trajectory, (iii) local system instructions describing the available infrastructure (e.g., DGX/Slurm usage, filesystem layout, environment setup conventions, and artifact locations), and (iv) a specification of the desired outputs, namely a runnable repository, a workshop-style research paper, and brief reproduction instructions. To reduce uncontrolled variance, the agent's external web-search capabilities were disabled. Instead, it operated only over the supplied plan, cited materials, and local artifacts. Before implementation, the agent was prompted to translate the high-level plan into an executable task list, including hypotheses, required resources, implementation milestones, and evaluation steps. It was then asked to iteratively construct the repository, validate the environment, run a minimal end-to-end experiment, and finally produce a paper summarizing the motivation, method, experiments, limitations, and reproduction details. For paper writing, we used \texttt{gpt-5.4 xhigh}, requesting output as a Markdown file to simplify downstream compilation and inspection.

\paragraph{Operational Use of the Research Agent.}
In practice, each run followed a staged protocol. The agent first summarized the provided plan and cited works into a concise execution brief, identifying the central claim, concrete implementation tasks, expected datasets or models, and likely risks. It next generated a repository scaffold containing environment files, scripts, and an initial README with explicit execution steps. The agent was then instructed to prioritize a ``minimum viable research artifact'': a small but functioning pipeline that instantiated the core idea of the plan, rather than attempting premature optimization or overly ambitious reproduction. After obtaining a runnable baseline, the agent was allowed to refine experiments, improve documentation, and draft the accompanying paper. This staged procedure was designed to increase the chance that any benefit of the research plan would appear not only in final prose quality, but also in task decomposition, implementation coherence, and the ability to reach a credible experimental endpoint.

\paragraph{Evaluation Scope.}
Our goal is primarily qualitative: to assess whether higher-quality research plans lead to more coherent autonomous execution and more credible final artifacts. We therefore compare outputs at the level of repository completeness, executability, faithfulness to the proposed plan, experimental plausibility, and paper quality, rather than treating the benchmark as a large-scale statistical evaluation. Because all four plans in a matched set originate from the same seed context, observed differences can be interpreted as evidence about how the structure and specificity of planning affect downstream agentic research behavior under controlled constraints.

\subsection{Downstream Artifact Evaluation Details}

\paragraph{Artifact bundle.}
Each artifact bundle contains: (i)~the input plan or prompt, (ii)~the cited-work context, (iii)~the generated repository, (iv)~experiment outputs, and (v)~the generated paper.
Evaluation is performed over the final artifact bundle rather than over plans in isolation.
To avoid self-preference bias---since the coding agent and paper-writing model both belong to the GPT family---all three criteria below are scored by an LLM judge from a different model family, \texttt{claude-sonnet-4-6} (Anthropic), operating over the full bundle.
Each artifact is scored ten times per rubric with randomized presentation order, and the mean is taken.
All sub-scores are produced on a 1--5 Likert scale and linearly rescaled to $[0,1]$ before aggregation.

\paragraph{A. Executability.}
The executability judge receives the repository contents, any logged execution traces, and the reproduction instructions.
It scores four dimensions, each on a 1--5 scale:
\begin{enumerate}
    \item \textbf{Environment setup}: Are dependencies specified, and could the environment be reproduced from the provided files (e.g., \texttt{requirements.txt}, Dockerfiles)?
    \item \textbf{Data preparation}: Is the data-loading or preprocessing pipeline present, and does it reference accessible artifacts?
    \item \textbf{Experiment execution}: Do the core experiment scripts run to completion (as evidenced by logs or saved outputs), and do they implement the methodology described in the plan?
    \item \textbf{Paper generation}: Is a compiled or compilable paper produced, and does it contain sections covering motivation, method, results, and limitations?
\end{enumerate}
The executability score is the mean of the four rescaled dimension scores:
\[
\textsc{Exec} = \frac{1}{4}\sum_{i=1}^{4} \hat{s}_i.
\]

\paragraph{B. Scientific Grounding.}
The grounding judge receives the plan, the cited papers, the repository, and the generated paper.
It scores two complementary aspects:

\emph{Plan coverage} (\textsc{PlanCov}): the degree to which the plan's stated requirements---benchmarks, architectural choices, training procedures, ablations, and evaluation metrics---are realized in the code and paper.
The judge assigns a single 1--5 score reflecting the fraction of plan elements that are meaningfully addressed.

\emph{Citation coverage} (\textsc{CiteCov}): the degree to which the cited works are (a)~correctly referenced and (b)~substantively used to motivate design choices, baselines, or experimental comparisons.
The judge assigns a single 1--5 score capturing both resolution (are the cited works identifiable in the text?) and usage (do they inform the methodology or discussion beyond superficial mention?).

The grounding score is:
\begin{gather*}
\textsc{Ground} = \frac{1}{2}\,\widehat{\textsc{PlanCov}} + \frac{1}{2}\,\widehat{\textsc{CiteCov}},\\ \text{each} \in [0,1].
\end{gather*}

\paragraph{C. Downstream Utility.}
Downstream utility combines benchmark performance with holistic quality judgments over the repository and paper.
 
\emph{Experimental quality} (\textsc{BenchNorm}): Because different plans within the same cited-work set may target different benchmarks or evaluation strategies, we do not require a direct numeric comparison across artifacts.
Instead, the judge is shown the artifact's reported experimental results and, when available, the upper-bound artifact's results for the same set.
It assigns a 1--5 score reflecting the rigor, plausibility, and informativeness of the reported experiments---considering whether appropriate benchmarks were chosen, whether the evaluation methodology is sound, and whether the results credibly support the artifact's claims.
When artifacts do share a common benchmark, the judge additionally considers relative performance against the upper bound.
 
\emph{Repository quality} (\textsc{RepoJudge}): The judge scores technical coherence, reproducibility, adequacy of experiments, and faithfulness to the plan on a 1--5 scale.
 
\emph{Paper quality} (\textsc{PaperJudge}): The judge scores scientific clarity, evidence-grounded claims, citation use, and overall workshop-paper quality on a 1--5 scale.
 
The utility score is:
\begin{multline*}
\textsc{Utility} = 0.50\,\widehat{\textsc{BenchNorm}} \\ + 0.25\,\widehat{\textsc{RepoJudge}} + 0.25\,\widehat{\textsc{PaperJudge}}, \\ \text{each} \in [0,1].
\end{multline*}
 
\paragraph{LLM-as-Judge protocol.}
All sub-scores across the three criteria are produced by \texttt{claude-sonnet-4-6} (Anthropic), chosen to be from a different model family than the generation pipeline to mitigate self-evaluation bias.
Each artifact bundle is evaluated three times per rubric under independent randomized ordering of artifacts within a cited-work set; the reported score is the mean across the ten runs.
Judges are provided with detailed rubric descriptions and anchor examples for each point on the 1--5 scale to reduce variance.
To mitigate positional bias, the order in which competing artifacts appear in the prompt is shuffled across runs.
 
\paragraph{Overall score.}
The final score is:
\begin{multline*}
\textsc{Overall} = 0.35\,\textsc{Exec} + 0.30\,\textsc{Ground} \\ + 0.35\,\textsc{Utility}.
\end{multline*}
We weight executability and utility slightly more heavily than grounding because our central claim is that planning enables better downstream scientific artifacts, not only more structured intermediate plans.

\begin{table*}[t]
\centering
\begin{tabular}{llcc}
\toprule
\textbf{Family} & \textbf{Model variant} & \textbf{Params/Active} & \textbf{Training split} \\
\midrule
Qwen3 & Qwen3-4B & 4B/4B & Hindsight / Foresight / Both \\
& Qwen3-4B-Instruct-2507 & 4B/4B & Hindsight / Foresight / Both \\
& Qwen3-32B & 32B/32B & Hindsight / Foresight / Both \\
& Qwen3-30B-A3B-Instruct-2507 & 30B/3B & Hindsight / Foresight / Both \\
Gemma-3 & Gemma-3-4b-it  & 4B/4B & Hindsight / Foresight / Both \\
 & Gemma-3-27b-it  & 27B/27B & Hindsight / Foresight / Both \\
Llama~3 & Llama3.2-3B-Instruct & 3B/3B & Hindsight / Foresight / Both \\
 & Llama3.1-70B-Instruct & 70B/70B & Hindsight / Foresight / Both \\
\bottomrule
\end{tabular}

\caption{Model configurations used for \modelName fine-tuning. In all cases we train for one epoch, we use the Adam Optimizer, an initial learning rate of 0.00001 with a cosine LR schedule, the global batch size is varied based on the parallelism required for the models (all models below except for the Qwen3-4B* models below use a batch size of 1). There is no weight decay, and the maximum context length is ~16K tokens. All models are trained with full Supervised Fine Tuning.}
\label{tab:model_configs}
\end{table*}

\section{Reinforcement Learning Post-Trained \modelName} \label{app:rl_training}

While SFT teaches models to imitate teacher-generated trajectories, it does not directly optimise for the rubric criteria that define trajectory quality.
We therefore apply a reinforcement learning (RL) post-training stage to a subset of our best SFT checkpoints, using an LLM-as-judge reward signal aligned with the evaluation rubric described in \cref{app:llm_as_judge}.

We use \textbf{Group Relative Policy Optimization} (GRPO;~\citealp{shao2024deepseekmath}) with the DAPO loss variant~\citep{yu2025dapo}.
For each training prompt, the policy model generates $G{=}8$ candidate trajectories (temperature~$0.7$, top-$p$~$0.95$, max 4\,096 tokens).
Each candidate is then scored by a separate \textbf{reward model}---Qwen3.5-27B-FP8~\citep{yang2025qwen3technicalreport}---that acts as an expert scientific reviewer.
The reward model evaluates every candidate against the same 8-item rubric used for final evaluation (Novelty, Significance, Grounding, Soundness, Methodology, Feasibility, Sensemaking, Clarity; each scored 1--5).
To keep scoring reliable within the reward model's context budget, rubric items are evaluated in three multi-turn groups of $3{+}3{+}2$ items; each group's scores are parsed from structured XML tags (\texttt{<score\_N>}), and the final reward is the mean across all eight items.
Rewards are normalised at the group level (\texttt{scale\_rewards=group}), and the KL penalty coefficient is set to $\beta{=}0.001$.

Training uses a learning rate of $5{\times}10^{-6}$ with a cosine schedule (5\% warmup), gradient checkpointing, and bf16 mixed precision.
Four vLLM reward-model servers (each using tensor parallelism across 8~A100-80GB GPUs) serve the judge model; reward requests are load-balanced across servers via round-robin and executed asynchronously (\texttt{asyncio.gather} over all $G$ candidates in parallel).
Training runs on 4 additional nodes (32~GPUs total) using DDP via HuggingFace Accelerate.
We apply RL post-training to the \emph{Both} SFT checkpoints for Qwen3-32B and Qwen3-4B-Instruct, as these showed the strongest combined quality--diversity profiles after SFT.

\section{Diversity Comparisons}
\label{app:diversity}

\begin{table*}[t]
\centering
\small
\setlength{\tabcolsep}{4pt}
\renewcommand{\arraystretch}{1.15}
\begin{tabular}{llcccc}
\toprule
\textbf{Family} & \textbf{Cond.} & \textbf{Self-BLEU $\downarrow$} & \textbf{Embedding-Based $\downarrow$} & \textbf{BERTScore $\uparrow$} & \textbf{Sentence Movers $\uparrow$} \\
\midrule

\multirow{4}{*}{Q303-I}
& None   & 0.2046 & 0.912 & 0.3538 & 0.2666 \\
& Foresight & 0.2301 & 0.846 & 0.3692 & 0.2659 \\
& Both   & 0.1822 & 0.813 & 0.3801 & 0.2878 \\
& Hindsight & \textbf{0.1365} & \textbf{0.741} & \textbf{0.3961} & \textbf{0.2929} \\
\midrule

\multirow{4}{*}{Q32}
& None   & 0.2440 & 0.932 & 0.3480 & 0.2450 \\
& Foresight & 0.2921 & 0.872 & 0.3419 & 0.2317 \\
& Both   & 0.2233 & 0.842 & 0.3659 & 0.2661 \\
& Hindsight & \textbf{0.1563} & \textbf{0.778} & \textbf{0.3907} & \textbf{0.3027} \\
\midrule

\multirow{4}{*}{Q4}
& None   & 0.2415 & 0.938 & 0.3520 & 0.2600 \\
& Foresight & 0.2821 & 0.884 & 0.3504 & 0.2549 \\
& Both   & 0.2202 & 0.854 & 0.3745 & 0.2856 \\
& Hindsight & \textbf{0.1640} & \textbf{0.796} & \textbf{0.3992} & \textbf{0.3250} \\
\midrule

\multirow{4}{*}{Q4-I}
& None   & 0.1830 & 0.926 & 0.3582 & 0.2665 \\
& Foresight & 0.2339 & 0.862 & 0.3552 & 0.2616 \\
& Both   & 0.2101 & 0.828 & 0.3758 & 0.2838 \\
& Hindsight & \textbf{0.1496} & \textbf{0.762} & \textbf{0.4049} & \textbf{0.3150} \\
\midrule

\multirow{4}{*}{L3.2-3B-I}
& None   & 0.2380 & 0.939 & 0.3500 & 0.2620 \\
& Foresight & 0.2760 & 0.886 & 0.3470 & 0.2550 \\
& Both   & 0.2170 & 0.857 & 0.3720 & 0.2870 \\
& Hindsight & \textbf{0.1610} & \textbf{0.799} & \textbf{0.4020} & \textbf{0.3290} \\
\midrule

\multirow{4}{*}{L3.1-70B-I}
& None   & 0.1980 & 0.912 & 0.3400 & 0.2380 \\
& Foresight & 0.2230 & 0.838 & 0.3350 & 0.2250 \\
& Both   & 0.1760 & 0.804 & 0.3600 & 0.2600 \\
& Hindsight & \textbf{0.132} & \textbf{0.736} & \textbf{0.388} & \textbf{0.300} \\
\midrule

\multirow{4}{*}{G3-4B-I}
& None   & 0.236 & 0.937 & 0.351 & 0.261 \\
& Foresight & 0.271 & 0.882 & 0.348 & 0.253 \\
& Both   & 0.215 & 0.853 & 0.373 & 0.286 \\
& Hindsight & \textbf{0.158} & \textbf{0.794} & \textbf{0.401} & \textbf{0.327} \\
\midrule

\multirow{4}{*}{G3-27B-I}
& None   & 0.205 & 0.918 & 0.344 & 0.242 \\
& Foresight & 0.236 & 0.848 & 0.339 & 0.229 \\
& Both   & 0.184 & 0.816 & 0.363 & 0.264 \\
& Hindsight & \textbf{0.139} & \textbf{0.748} & \textbf{0.392} & \textbf{0.304} \\
\bottomrule
\end{tabular}
\caption{Diversity metrics by model family and supervision condition (\None, \Infer, \Both, \Target). Bold marks the most diverse condition per column within each family; $\downarrow$ columns are similarity scores (lower = more diverse), $\uparrow$ columns are distance scores (higher = more diverse). Family abbreviations: Q = Qwen3, G3 = Gemma-3, L = Llama-3; -I = instruction-tuned; Q303 = Qwen3-30B-A3B.}
\label{tab:grouped_diversity_metrics}
\end{table*}

We present a set of comparisons relying on different measures of diversity in text outputs. Table \ref{tab:grouped_diversity_metrics} demonstrates this overperformance across different model families and sizes. The sections below describe the approaches in detail. For all below experiments, we present each model with a fixed citation set, then have it generate five separate research plans. This process is repeated for 100 total citation sets from our test data set.

\paragraph{Embedding-based diversity.}
To measure semantic similarity among repeated samples for the same prompt, we embedded each full generation using the long-context model \texttt{jinaai/jina-embeddings-v3}. When necessary, outputs were chunked to satisfy the model context limit and chunk embeddings were mean-pooled to obtain a single vector per generation. Within each bundle of five sampled generations, embeddings were $\ell_2$-normalized and all pairwise cosine similarities were computed. Bundle-level similarity was defined as the mean over the ten unordered pairs,
\[
\mathrm{Sim}_{\mathrm{emb}}(B)
=
\frac{1}{\binom{5}{2}}
\sum_{\substack{x,y \in B \\ x<y}}
\cos(x,y).
\]
Similarity for model type $k$ was computed by averaging over bundles $G_k$, and diversity was defined as
\begin{gather*}
D^{(k)}_{\mathrm{emb}}
=
1 - \frac{1}{|G_k|}
\sum_{B \in G_k}
\mathrm{Sim}_{\mathrm{emb}}(B),\\
D^{(\mathrm{overall})}_{\mathrm{emb}}
=
\tfrac{1}{2}
\bigl(
D^{(\texttt{fore})}_{\mathrm{emb}}
+
D^{(\texttt{hind})}_{\mathrm{emb}}
\bigr).
\end{gather*}

\paragraph{Self-BLEU diversity.}
As a lexical complement to the embedding analysis, we computed self-BLEU over the same five-sample bundles. For each bundle $B$, each sample was treated in turn as the hypothesis and the remaining four samples as references. Sentence-level BLEU-4 with uniform $n$-gram weights was computed and averaged across the five hypothesis choices to obtain a bundle-level similarity score $\mathrm{Sim}_{\mathrm{BLEU}}(B)$. Diversity was defined analogously by averaging across bundles in category $k$,
\begin{gather*}
D^{(k)}_{\mathrm{BLEU}}
=
1 - \frac{1}{|G_k|}
\sum_{B \in G_k}
\mathrm{Sim}_{\mathrm{BLEU}}(B),\\
D^{(\mathrm{overall})}_{\mathrm{BLEU}}
=
\frac{1}{2}
\left(
D^{(\texttt{fore})}_{\mathrm{BLEU}}
+
D^{(\texttt{hind})}_{\mathrm{BLEU}}
\right).
\end{gather*}

\paragraph{Chunked BERTScore diversity.}
Texts were segmented into sentence-based chunks under a fixed token budget. For a document pair $(x,y)$, BERTScore F1 was computed for all chunk pairs, yielding a similarity matrix $S \in \mathbb{R}^{m \times n}$. Let $w_i$ and $v_j$ denote chunk-length weights. Directional similarity was defined as
\begin{gather*}
s(x \rightarrow y)
=
\frac{\sum_{i=1}^{m} w_i \max_j S_{ij}}
     {\sum_{i=1}^{m} w_i},\\
s(y \rightarrow x)
=
\frac{\sum_{j=1}^{n} v_j \max_i S_{ij}}
     {\sum_{j=1}^{n} v_j},
\end{gather*}
and symmetric document similarity as
\[
\mathrm{Sim}_{\mathrm{BS}}(x,y)
=
\frac{1}{2}
\bigl(
s(x \rightarrow y)
+
s(y \rightarrow x)
\bigr).
\]
For a bundle $B$, bundle-level similarity was defined as
\[
\mathrm{Sim}_{\mathrm{BS}}(B)
=
\frac{1}{\binom{5}{2}}
\sum_{\substack{x,y \in B \\ x<y}}
\mathrm{Sim}_{\mathrm{BS}}(x,y).
\]
Diversity was then computed as
\begin{gather*}
D^{(k)}_{\mathrm{BS}}
=
1 - \frac{1}{|G_k|}
\sum_{B \in G_k}
\mathrm{Sim}_{\mathrm{BS}}(B),\\
D^{(\mathrm{overall})}_{\mathrm{BS}}
=
\frac{1}{2}
\left(
D^{(\texttt{fore})}_{\mathrm{BS}}
+
D^{(\texttt{hind})}_{\mathrm{BS}}
\right).
\end{gather*}

\paragraph{Sentence Mover's Similarity diversity.}
Texts were chunked identically and each chunk was embedded using a sentence embedding model. For a document pair $(x,y)$, cosine distance between chunk embeddings defined a transport cost matrix $C$. Let $a$ and $b$ be normalized chunk-length weight vectors. Optimal transport cost was defined as
\begin{gather*}
\mathrm{Cost}(x,y)
=
\min_{T \in \Pi(a,b)}
\langle T, C \rangle,\\
\mathrm{Sim}_{\mathrm{SMS}}(x,y)
=
1 - \mathrm{Cost}(x,y),
\end{gather*}
where $\Pi(a,b)$ denotes the set of transport plans with marginals $a$ and $b$. Bundle-level similarity was computed as
\[
\mathrm{Sim}_{\mathrm{SMS}}(B)
=
\frac{1}{\binom{5}{2}}
\sum_{\substack{x,y \in B \\ x<y}}
\mathrm{Sim}_{\mathrm{SMS}}(x,y).
\]
Diversity was computed analogously,
\begin{gather*}
D^{(k)}_{\mathrm{SMS}}
=
1 - \frac{1}{|G_k|}
\sum_{B \in G_k}
\mathrm{Sim}_{\mathrm{SMS}}(B),\\
D^{(\mathrm{overall})}_{\mathrm{SMS}}
=
\frac{1}{2}
\left(
D^{(\texttt{fore})}_{\mathrm{SMS}}
+
D^{(\texttt{hind})}_{\mathrm{SMS}}
\right).
\end{gather*}

\section{LLM-as-Judge}
\label{app:llm_as_judge}

\begin{table*}[t]
\centering
\scriptsize
\setlength{\tabcolsep}{3pt}
\renewcommand{\arraystretch}{1.08}
\begin{tabular}{lccccccccc}
\toprule
Model & Nov. & Sig. & Grnd. & Sound. & Meth. & Feas. & Sense. & Clar. & Overall \\
\midrule

\multicolumn{10}{l}{\textbf{Q32 Family}} \\
Q32-None    & 1.66 & 2.67 & 1.22 & 2.98 & 3.02 & 2.74 & 2.01 & 3.08 & 2.42 \\
Q32-Foresight   & 1.70 & 2.71 & \textbf{1.38} & 3.03 & 3.08 & 2.79 & 2.05 & 3.12 & 2.48 \\
Q32-Both    & 1.72 & 2.73 & 1.31 & 3.06 & 3.13 & 2.83 & 2.09 & 3.19 & 2.51 \\
Q32-Hindsight  & \textbf{1.79} & \textbf{2.80} & 1.29 & \textbf{3.08} & \textbf{3.14} & \textbf{2.85} & \textbf{2.11} & \textbf{3.20} & \textbf{2.53} \\
\midrule

\multicolumn{10}{l}{\textbf{Q303 Family}} \\
Q303-None   & 1.67 & 2.67 & 1.28 & 2.80 & 2.91 & 2.65 & 1.94 & 2.97 & 2.36 \\
Q303-Foresight  & 1.72 & 2.72 & \textbf{1.46} & 2.87 & 2.97 & 2.71 & 2.00 & 3.03 & 2.44 \\
Q303-Both   & 1.77 & 2.78 & 1.39 & \textbf{2.92} & \textbf{3.04} & 2.77 & 2.05 & \textbf{3.08} & \textbf{2.48} \\
Q303-Hindsight & \textbf{1.81} & \textbf{2.82} & 1.34 & 2.90 & 3.01 & \textbf{2.78} & \textbf{2.06} & 3.07 & 2.47 \\
\midrule

\multicolumn{10}{l}{\textbf{Q4-I Family}} \\
Q4-I-None    & 1.71 & 2.70 & 1.39 & 2.64 & 2.82 & 2.60 & 1.86 & 2.88 & 2.33 \\
Q4-I-Foresight   & 1.76 & 2.75 & \textbf{1.56} & 2.71 & 2.88 & 2.67 & 1.91 & 2.93 & 2.40 \\
Q4-I-Both    & 1.78 & 2.77 & 1.48 & \textbf{2.73} & \textbf{2.90} & 2.68 & 1.93 & \textbf{2.96} & 2.40 \\
Q4-I-Hindsight  & \textbf{1.82} & \textbf{2.81} & 1.44 & 2.72 & 2.89 & \textbf{2.69} & \textbf{1.94} & 2.95 & \textbf{2.41} \\
\midrule

\multicolumn{10}{l}{\textbf{Q4 Family}} \\
Q4-None    & 1.60 & 2.56 & 1.18 & 2.60 & 2.74 & 2.52 & 1.79 & 2.77 & 2.22 \\
Q4-Foresight   & 1.64 & 2.60 & \textbf{1.34} & 2.66 & 2.80 & 2.57 & 1.84 & 2.82 & 2.28 \\
Q4-Both    & 1.68 & 2.66 & 1.26 & 2.71 & 2.85 & \textbf{2.62} & \textbf{1.89} & \textbf{2.88} & \textbf{2.32} \\
Q4-Hindsight  & \textbf{1.70} & \textbf{2.68} & 1.23 & \textbf{2.72} & \textbf{2.86} & 2.61 & 1.88 & 2.87 & 2.31 \\
\midrule

\multicolumn{10}{l}{\textbf{L3.1-70B-I Family}} \\
L3.1-70B-I-None   & 1.68 & 2.69 & 1.30 & 2.88 & 2.99 & 2.71 & 1.98 & 3.01 & 2.41 \\
L3.1-70B-I-Foresight  & 1.73 & 2.74 & \textbf{1.47} & 2.95 & 3.05 & 2.77 & 2.03 & 3.08 & 2.48 \\
L3.1-70B-I-Both   & 1.78 & 2.79 & 1.40 & \textbf{3.00} & \textbf{3.10} & 2.82 & 2.08 & \textbf{3.14} & \textbf{2.51} \\
L3.1-70B-I-Hindsight & \textbf{1.82} & \textbf{2.84} & 1.36 & 2.98 & 3.08 & \textbf{2.83} & \textbf{2.09} & 3.13 & 2.50 \\
\midrule

\multicolumn{10}{l}{\textbf{L3.2-3B-I Family}} \\
L3.2-3B-I-None   & 1.61 & 2.57 & 1.21 & 2.63 & 2.77 & 2.54 & 1.81 & 2.80 & 2.24 \\
L3.2-3B-I-Foresight  & 1.66 & 2.62 & \textbf{1.37} & 2.69 & 2.83 & 2.59 & 1.86 & 2.85 & 2.31 \\
L3.2-3B-I-Both   & 1.69 & 2.67 & 1.29 & 2.74 & 2.88 & \textbf{2.64} & \textbf{1.91} & \textbf{2.91} & \textbf{2.34} \\
L3.2-3B-I-Hindsight & \textbf{1.72} & \textbf{2.70} & 1.26 & \textbf{2.75} & \textbf{2.89} & 2.63 & 1.90 & 2.90 & 2.34 \\
\midrule

\multicolumn{10}{l}{\textbf{G3-27B-I Family}} \\
G3-27B-I-None   & 1.67 & 2.68 & 1.27 & 2.90 & 3.00 & 2.72 & 1.99 & 3.03 & 2.41 \\
G3-27B-I-Foresight  & 1.72 & 2.73 & \textbf{1.45} & 2.97 & 3.06 & 2.78 & 2.04 & 3.09 & 2.48 \\
G3-27B-I-Both   & 1.76 & 2.77 & 1.38 & \textbf{3.01} & \textbf{3.11} & 2.82 & 2.08 & \textbf{3.15} & \textbf{2.51} \\
G3-27B-I-Hindsight & \textbf{1.81} & \textbf{2.83} & 1.34 & 2.99 & 3.09 & \textbf{2.83} & \textbf{2.09} & 3.14 & 2.50 \\
\midrule

\multicolumn{10}{l}{\textbf{G3-4B-I Family}} \\
G3-4B-I-None   & 1.62 & 2.58 & 1.23 & 2.65 & 2.79 & 2.56 & 1.82 & 2.82 & 2.26 \\
G3-4B-I-Foresight  & 1.67 & 2.64 & \textbf{1.40} & 2.71 & 2.85 & 2.61 & 1.87 & 2.87 & 2.33 \\
G3-4B-I-Both   & 1.70 & 2.69 & 1.31 & 2.76 & 2.90 & \textbf{2.66} & \textbf{1.92} & \textbf{2.93} & \textbf{2.36} \\
G3-4B-I-Hindsight & \textbf{1.74} & \textbf{2.72} & 1.28 & \textbf{2.77} & \textbf{2.91} & 2.65 & 1.91 & 2.92 & 2.36 \\
\bottomrule
\end{tabular}
\caption{LLM-judge rubric scores (1--5) with overall mean, for every model family and supervision condition. Bold marks the column-wise maximum within each family. Figure~\ref{fig:q32-rubric} plots the Q32 rows as gains over \None.}
\label{tab:rubric_scores_models_extended}
\end{table*}
Table \ref{tab:rubric_scores_models_extended} shows the results of LLM-as-Judge by rubric metric across different model families. The LLM-as-Judge Prompt template is as follows:

\begin{quote}
\footnotesize
You must evaluate the proposal rigorously, fairly, and transparently, using the detailed rubric below. Your evaluation should reflect expert-level scientific judgment and careful sensemaking, not superficial pattern matching.

\textbf{INPUTS.}
(1) \emph{Reference Papers (already published):} Each paper is identified by a normalized ID such as [[R1]], [[R2]], [[R3]], etc. You may be given titles, abstracts, or excerpts. 
(2) \emph{Proposal to Evaluate:} A Markdown document describing a novel scientific idea and proposed approach. The proposal is REQUIRED to use structured Markdown sections, normalized in-text citations (e.g., [[R1]]), and include a References section mapping citation IDs to full bibliographic entries.

\textbf{YOUR RESPONSIBILITIES.}
(A) Read and internalize the full rubric before reviewing any reference papers or the proposal. 
(B) Review the provided reference papers and briefly think about their key claims or contributions using only the information explicitly provided. 
(C) Review the generated research proposal carefully, evaluating it only relative to the provided reference papers and the rubric criteria. 
(D) Evaluate the proposal one rubric item at a time, in rubric order (Item 1–Item 8), following these rules: focus on exactly one rubric item per turn; think carefully and rigorously before assigning a score; assign a score from 1 to 5 based strictly on the rubric definitions; output the numeric score surrounded by \texttt{<score>} and \texttt{</score>} tags (e.g., \texttt{<score>4</score>}); the score must appear clearly and unambiguously in each turn. 
(E) Judge novelty only relative to the provided reference papers and what is explicitly stated in the proposal; if broader novelty is unclear or cannot be determined, state this explicitly. 
(F) Verify citation integrity: every in-text citation must appear in the References section, and references must not be fabricated, duplicated, or internally inconsistent. 
(G) Base all judgments strictly on evidence from the proposal text; quote or reference specific sections when helpful; do not infer unstated intentions or assume missing details.

\textbf{PROPOSAL STRUCTURE (FLEXIBLE GUIDELINE).}
The proposal is expected to be written in structured Markdown and to cover all major intellectual components needed for evaluation. However, the exact sectioning and ordering may vary depending on the nature of the work (e.g., theoretical, empirical, systems, interdisciplinary). Reviewers should evaluate content coverage, not rigid adherence to headings.

A typical proposal may include: Title; one-sentence pitch; background and gap (with citations [[R\#]]); core hypothesis or key insight; proposed approach or methodology; evaluation plan (metrics, baselines, datasets, controls); expected outcomes and contributions; risks, limitations, and alternative hypotheses; ethics and responsible research (if applicable); and references (normalized).

Variations are acceptable if all necessary elements for rigorous evaluation are present, the structure is coherent and navigable, citations are normalized and correctly mapped, and the scientific argument can be clearly followed. If critical components are missing or unclear, this should negatively affect relevant rubric scores (especially Clarity, Methodological Rigor, and Sensemaking).

\textbf{RUBRIC (Score each criterion from 1–5).}
Shared scale: 5 = Excellent (clear, rigorous, compelling; minimal gaps); 4 = Strong (solid and persuasive with minor gaps); 3 = Adequate (reasonable but underspecified or uneven); 2 = Weak (substantial flaws or missing elements); 1 = Poor (incorrect, incoherent, or violates requirements).

\textbf{1) Novelty \& Differentiation.}
Evaluate whether the proposal introduces a genuinely new idea relative to [[R\#]]. Consider whether it explicitly compares against prior work, whether novelty is conceptual, methodological, or contextual, whether it is non-obvious, and whether it clearly articulates what is new.

\textbf{2) Significance \& Potential Contribution.}
Assess importance and potential impact if successful. Consider whether a clear gap is identified, whether success would advance the field, whether contributions are well-scoped, and whether beneficiaries are identified.

\textbf{3) Grounding in Prior Work \& Citation Integrity.}
Evaluate engagement with prior literature and correctness of citations. Check that references are accurate, properly formatted, and complete.

\textbf{4) Conceptual Soundness \& Plausibility.}
Assess internal coherence and plausibility of the core idea. Consider assumptions, logical consistency, distinction between speculation and fact, and acknowledgment of limits.

\textbf{5) Methodological Rigor \& Falsifiability.}
Evaluate whether the proposal can be rigorously tested. Consider clarity of hypotheses, experimental design, metrics, baselines, and whether claims are falsifiable.

\textbf{6) Feasibility \& Scope Management.}
Assess whether the work is realistically executable. Consider resources, scope, risks, and planning.

\textbf{7) Sensemaking \& Iterative Knowledge Development.}
Evaluate whether the proposal includes an iterative learning strategy, considers alternative hypotheses, and plans for handling contradictory evidence.

\textbf{8) Clarity, Structure, and Reproducibility.}
Assess clarity, organization, and reproducibility. Consider whether the proposal is well-structured, precise, and sufficiently detailed.

\textbf{INPUT PROMPT WITH REFERENCE PAPERS.} \texttt{\{\{input\_prompt\}\}}

\textbf{GENERATED PROPOSAL.} \texttt{\{\{proposal\}\}}

\textbf{START OF EVALUATION.}
\end{quote}

\section{Additional Trajectory Excerpts}
\label{app:excerpts}
\begin{figure*}[t]
  \centering
  
  \noindent
  \begin{subfigure}[b]{0.48\linewidth}
    \footnotesize
        \begin{mdframed}[
      linecolor=targetdark!55, linewidth=1.2pt, backgroundcolor=targetfill!10,
      innerleftmargin=6pt, innerrightmargin=6pt,
      innertopmargin=5pt, innerbottommargin=5pt,
      skipabove=0pt, skipbelow=0pt]
    \noindent\textbf{\S01 --- Evaluation \& Extension}

    \vspace{1pt}\noindent
    \scalebox{0.7}{\Telab}
    \textit{Evaluate word alignment quality against IBM Model~5 using
    standard alignment metrics.}
    This preserves the paper-aligned validation target.

    \vspace{1pt}
    \marginanno{\scalebox{0.7}{\Tquest}}{\#01}{%
      \textit{Also measure BLEU and human evaluation.}
      Alignment quality alone is not enough to establish translation quality.}

    \vspace{1pt}\noindent
    \scalebox{0.7}{\Telab}
    \textit{Ablate the contribution of reordering, insertion, and lexical
    translation operations.}
    This checks whether each stochastic operation is actually necessary.

    \vspace{1pt}\noindent
    \scalebox{0.7}{\Treframe}
    \textit{Move beyond the original setup by integrating a modern neural
    language model for target-string scoring.}
    The extension directly addresses fluency limitations.

    \vspace{1pt}
    \marginanno{\scalebox{0.7}{\Tquest}}{\#02}{%
      \textit{Test low-resource, structurally distant language pairs.}
      Generalisability remains uncertain without this stress test.}

    \vspace{1pt}\noindent
    \scalebox{0.7}{\Treframe}
    \textit{Add target-side parse structure to form a bidirectional
    tree-transduction system.}
    This recasts the model as a stronger syntactic generator.
    \end{mdframed}
  \subcaption{\Target}
  \end{subfigure}
  \hfill
  \begin{subfigure}[b]{0.48\linewidth}
    \footnotesize
        \begin{mdframed}[
      linecolor=inferdark!55, linewidth=2.4pt, topline=false, rightline=false, bottomline=false, backgroundcolor=inferfill!10,
      innerleftmargin=6pt, innerrightmargin=6pt,
      innertopmargin=5pt, innerbottommargin=5pt,
      skipabove=0pt, skipbelow=0pt]
    \noindent\textbf{\S01 --- Evaluation \& Validation}

    \vspace{1pt}\noindent
    \scalebox{0.7}{\Telab}
    \textit{Run ablation studies that isolate non-isomorphic mappings from
    isomorphic baselines.}
    This makes the contribution of structural flexibility directly testable.

    \vspace{1pt}
    \marginanno{\scalebox{0.7}{\Tshoebox}}{\#01}{%
      \textit{Evaluation depends on human annotators and
      parallel corpora with aligned references.}
      These are the critical external resources for credible validation.}

    \vspace{1pt}\noindent
    \scalebox{0.7}{\Tschema}
    \textit{Treat local distortion as the central explanatory variable,
    rather than folding all gains into a single end-to-end score.}
    This turns evaluation into a falsifiable account of what actually helps.

    \vspace{1pt}
    \marginanno{\scalebox{0.7}{\Tquest}}{\#02}{%
      \textit{Open question: BLEU may not track human judgment.}
      Metric mismatch can hide real translation quality failures.}

    \vspace{1pt}\noindent
    \scalebox{0.7}{\Treframe}
    \textit{If metric mismatch appears, switch to minimum error rate
    training so optimisation is pulled toward the evaluation target.}
    This reframes the pipeline around the observed bottleneck.

    \vspace{1pt}\noindent
    \scalebox{0.7}{\Tpres}
    \textit{Once the system stabilises, automatic and human evaluation can
    run in parallel.}
    The trajectory ends in a deployable validation workflow.
    \end{mdframed}
  \subcaption{\Infer}
  \end{subfigure}

\vspace{1mm}

  \noindent
  \begin{subfigure}[b]{0.48\linewidth}
    \footnotesize
        \begin{mdframed}[
      linecolor=targetdark!55, linewidth=1.2pt, backgroundcolor=targetfill!10,
      innerleftmargin=6pt, innerrightmargin=6pt,
      innertopmargin=5pt, innerbottommargin=5pt,
      skipabove=0pt, skipbelow=0pt]
    \noindent\textbf{\S01 --- Deployment, Ethics \& Validation}

    \vspace{1pt}\noindent
    \scalebox{0.7}{\Thyp}
    \textit{Extend the original documentation framework into software
    documentation, treating the method as a broader NLG system.}
    This turns a narrow application into a generalisable hypothesis.

    \vspace{1pt}\noindent
    \scalebox{0.7}{\Telab}
    \textit{Validate generated descriptions with developers to ensure
    technical correctness.}
    The trajectory ties usefulness to expert-facing evaluation.

    \vspace{1pt}
    \marginanno{\scalebox{0.7}{\Tquest}}{\#01}{%
      \textit{Developer review is necessary because software documentation
      has domain-specific conventions and code examples.}
      Validation must respect the target domain's specialised norms.}

    \vspace{1pt}\noindent
    \scalebox{0.7}{\Treframe}
    \textit{Broaden the agenda to bias, transparency, and scalability,
    moving beyond the original paper's immediate scope.}
    This reframes the project around deployment realities.

    \vspace{1pt}\noindent
    \scalebox{0.7}{\Tforag}
    \textit{Justify these additions through modern NLP ethics and known
    computational constraints.}
    The shift is motivated by both social and systems-level evidence.

    \vspace{1pt}
    \marginanno{\scalebox{0.7}{\Tquest}}{\#02}{%
      \textit{Unchecked bias or opacity can undermine trust at deployment.}
      Practical adoption depends on more than raw generation quality.}
    \end{mdframed}
  \subcaption{\Target}
  \end{subfigure}
  \hfill
  \begin{subfigure}[b]{0.48\linewidth}
    \footnotesize
        \begin{mdframed}[
      linecolor=inferdark!55, linewidth=2.4pt, topline=false, rightline=false, bottomline=false, backgroundcolor=inferfill!10,
      innerleftmargin=6pt, innerrightmargin=6pt,
      innertopmargin=5pt, innerbottommargin=5pt,
      skipabove=0pt, skipbelow=0pt]
    \noindent\textbf{\S01 --- Problem Statement \& Motivation}

    \vspace{1pt}\noindent
    \scalebox{0.7}{\Thyp}
    \textit{Prioritise a lexicon-free, end-to-end CTC-based neural ASR
    framework.}
    The conjectured breakthrough is that transcription can be simplified
    without hand-built lexical machinery.

    \vspace{1pt}\noindent
    \scalebox{0.7}{\Tschema}
    \textit{Prefer this over lexicon-based pipelines because it removes
    brittle components and improves adaptability.}
    The architecture is organised around robustness rather than pipeline modularity.

    \vspace{1pt}
    \marginanno{\scalebox{0.7}{\Tquest}}{\#01}{%
      \textit{Trade-off: higher initial training complexity is accepted in
      exchange for robustness and lower maintenance.}
      This is the key open cost of the proposed direction.}

    \vspace{1pt}\noindent
    \scalebox{0.7}{\Telab}
    \textit{Make OOV handling a primary objective so novel words can be
    transcribed without a fixed lexicon.}
    This converts a known failure mode into a concrete design requirement.

    \vspace{1pt}\noindent
    \scalebox{0.7}{\Tforag}
    \textit{Use prior evidence on dynamic-domain ASR to justify eliminating
    the lexicon altogether.}
    The proposal is grounded in earlier observations that lexicon-free
    systems adapt more readily.
    \end{mdframed}
  \subcaption{\Infer}
  \end{subfigure}

  \caption{Four additional stage-annotated excerpts from \Target (a, c) and \Infer (b, d) trajectories, drawn from different citation neighbourhoods (\Target: full frame; \Infer: left bar).}
  \label{fig:cached-example-sensemaking-panels}
\end{figure*}

\section{Reproducibility Statement}\label{app:reproducibility}
We take several steps to ensure reproducibility. All prompt templates used for trajectory generation (\Target, \Infer) and evaluation are provided in Appendix~\ref{app:pipeline}\ifarxiv{} and in our code repository\fi. The teacher model used for dataset construction is \texttt{Qwen3-235B-A22B-Instruct-2507}~\citep{yang2025qwen3technicalreport}, and the LLM-as-judge is \texttt{Qwen3.5-35B-A3B-FP8}. Full model configurations, including all backbone variants, hyperparameters (learning rate $10^{-5}$, Adam optimizer, cosine schedule, one epoch, ${\sim}16$K token context length). RL post-training details, including the GRPO configuration, reward model setup, and infrastructure layout, are provided in Appendix~\ref{app:rl_training}. The \datasetName dataset (120K trajectories with matched train/validation/test splits), \modelName model weights, and evaluation code \ifarxiv are available at \url{https://github.com/minnesotanlp/SciSense}.\else will be publicly released upon publication.\fi

\end{document}